\documentclass[a4paper,fleqn]{cas-sc}

\usepackage[authoryear]{natbib}
\usepackage{subcaption}
\usepackage{algorithm}
\usepackage[noend]{algpseudocode}
\usepackage{svg}
\def\tsc#1{\csdef{#1}{\textsc{\lowercase{#1}}\xspace}}
\tsc{WGM}
\tsc{QE}
\usepackage{svg}
\usepackage{xurl}
\usepackage{hyperref}
\usepackage{cleveref}
\usepackage{siunitx}
\usepackage{float}
\usepackage[section]{placeins}

\usepackage{enumitem}
\usepackage{booktabs}
\usepackage{array}

\usepackage{graphicx}
\usepackage{caption}
\usepackage{subcaption}
\usepackage{float}
\usepackage{booktabs}
\usepackage{threeparttable}

\usepackage{longtable}
\usepackage{soul}

\usepackage{pdflscape}
\usepackage{xltabular}

\usepackage{chngcntr}
\usepackage{placeins}   % for \FloatBarrier

\crefname{figure}{Figure}{Figures}       
\Crefname{figure}{Figure}{Figures}       
\crefname{section}{Section}{Sections}
\Crefname{section}{Section}{Sections}
\begin{document}
\let\WriteBookmarks\relax
\def\floatpagepagefraction{1}
\def\textpagefraction{.001}
%\shorttitle{Uncertainty-Aware Functional Behavior and Reliability Prediction for Angle Grinders in Circular Factories}    
\shorttitle{Uncertainty--Aware Functional Behavior Prediction and Material-Fatigue Assessment for Circular Factory Redeployment}    
\shortauthors{Afifi. et al.}  

%\title [mode = title]{Uncertainty-Aware Functional Behavior and Reliability Prediction for Angle Grinders in Circular Factories}  
\title [mode = title]{Uncertainty--Aware Functional Behavior Prediction and Material-Fatigue Assessment for Circular Factory}  

\tnotemark[1] 

\tnotetext[0]{This research was funded by the Deutsche Forschungsgemeinschaft (DFG, German Research Foundation) in the collaborative research center (CRC) 1574 “Circular Factory for the Perpetual Product” with the project ID 471687386. } 

%      orcid=0000-0000-0000-0000,
%\author[1]{Nehal Afifi}[type=editor,
%                      auid=000,bioid=1,
%                        prefix=,
%                        role=,
%                        orcid=0009-0002-7816-1595]
\author[1]{Nehal Afifi}[type=editor,
                        auid=000,bioid=1,
                        prefix=,
                        role=,
                        orcid=0009-0002-7816-1595]
%\author[1]{Nehal Afifi}%[<options>]
\cormark[1]
\fnmark[1]
\ead{nehal.afifi@kit.edu}
\ead[url]{https://www.ipek.kit.edu/21_11088.php}
%0009-0002-7816-1595
\credit{Conceptualization, Methodology, Investigation, Formal analysis, Visualization, Writing -- original draft, Writing -- review \& editing}
\affiliation[1]{organization={IPEK Institute of Product Engineering, Karlsruhe Institute of Technology (KIT)},%Department and Organization
            addressline={Kaiserstr. 10}, 
            city={Karlsruhe},
            postcode={76131}, 
            state={Baden Wuerttemberg},
            country={Germany}}

\author[2]{Mehdi Khabou} %% Author name
\credit{Conceptualization, Methodology, Investigation, Formal analysis, Visualization, Writing -- original draft, Writing -- review \& editing}
\author[1]{Victor Mas} %% Author name
\credit{Conceptualization, Writing -- original draft}
\author[1]{Jonas Hemmerich} %% Author name
\credit{Conceptualization, Writing -- original draft}
\author[1]{Patric Grauberger} %% Author n
\credit{Supervision, Writing -- review \& editing}
\author[2]{Stefan Dietrich} %% Author name
\credit{Supervision, Writing -- review \& editing}
\author[2,3]{Volker Schulze} %% Author name
\credit{Funding acquisition, Supervision, Writing -- review \& editing}
\author[1]{Sven Matthiesen} %% Author name
% Footnote of the second author
\fnmark[2]

% Email id of the second author
\ead{sven.matthiesen@kit.edu}

\ead[url]{https://www.ipek.kit.edu/21_425.php}
\credit{Supervision, Project administration, Conceptualization, Writing -- review \& editing}
\affiliation[2]{organization={IAM-WK Institute for Applied Materials – Materials Science and Engineering, Karlsruhe Institute of Technology (KIT)},%Department and Organization
            addressline={Engelbert-Arnold-Str. 4}, 
            city={Karlsruhe},
            postcode={76131}, 
            state={Baden Wuerttemberg},
            country={Germany}}

\affiliation[3]{organization={wbk Institute of Production Science, Karlsruhe Institute of Technology (KIT)},%Department and Organization
            addressline={Kaiserstr. 12}, 
            city={Karlsruhe},
            postcode={76131}, 
            state={Baden Wuerttemberg},
            country={Germany}}

\cortext[1]{Corresponding author}

\fntext[1]{}


\begin{abstract}
Returned products in circular factories re-enter production with heterogeneous degradation states, usage histories, and remaining capability. Reuse cannot be decided from the current inspection alone, because future function fulfillment and component integrity may evolve differently under the next service scenario. Existing PHM approaches support degradation prediction, but often target fixed operating conditions or isolated component benchmarks, while material-fatigue assessment is rarely linked to system-level functional prognosis. This paper addresses this gap for an angle grinder by combining uncertainty-aware functional prediction with component-level fatigue assessment in an instance-specific reliability workflow.
The proposed framework combines the current tool state with recent force--torque usage windows. A convolutional encoder extracts loading patterns from spindle forces and shaft torque, and an LSTM backbone predicts nine functional variables as Gaussian mean and variance estimates. In parallel, the same loading history is translated into output-shaft fatigue information through finite-element-supported stress reconstruction, S--N/Miner damage evaluation with Haibach extension, and Paris-law crack-growth analysis. A streaming replay algorithm consolidates both branches into functional and material reliability estimates and a system reliability indicator identifying the limiting assessed requirement.
Held-out tests show mean \(2\%\)-tolerance accuracy of 0.9652 across nine outputs. Thermal variables are predicted near-perfectly, while drive motor current and load speed remain the most demanding dynamic outputs, with \(R^2\) values of 0.9750 and 0.9924. Torque history is especially important for these variables, and the conventional LSTM outperforms GRU and xLSTM in the short-history setting. Reliability calibration is most informative for drive motor current, where predicted and observed exceedance probabilities agree closely. Material results show negligible Miner damage under nominal service stresses, but Paris-law analysis is sensitive to rare high-load events, reducing nominal shaft reusability from about 31 service phases to 3 under a 1.6 amplification of the upper stress tail. The integrated replay demonstrates a practical basis for instance-specific redeployment decisions while highlighting the need for more fatigue-critical and heterogeneous validation data.
\end{abstract}

% Use if a graphical abstract is present

\begin{highlights}
\item Conditional sequence learning is used to model the functional behavior of the used angle grinder.
\item Tool state and usage history are jointly exploited for future behavior prediction
\item Prescribed future usage windows enable scenario-conditioned reliability assessment.
\item An uncertainty-aware LSTM estimates both future behavior and predictive confidence.
\item Material-informed degradation indicators are linked to physical functional degradation
\item The material branch combines FE stress reconstruction, S--N/Miner damage, and Paris crack growth.
\item Threshold-based prediction enables reliability assessment.
\item Ablation studies quantify the effect of uncertainty modeling and input design.
\item Streaming replay consolidates functional and material reliability for system reliability assessment.
\item High-load events strongly reduce predicted output-shaft reuse potential.
\end{highlights}
\begin{graphicalabstract}
     \includegraphics[width=1.0\linewidth, trim=0.0cm 0.0cm 0cm 0cm, clip]{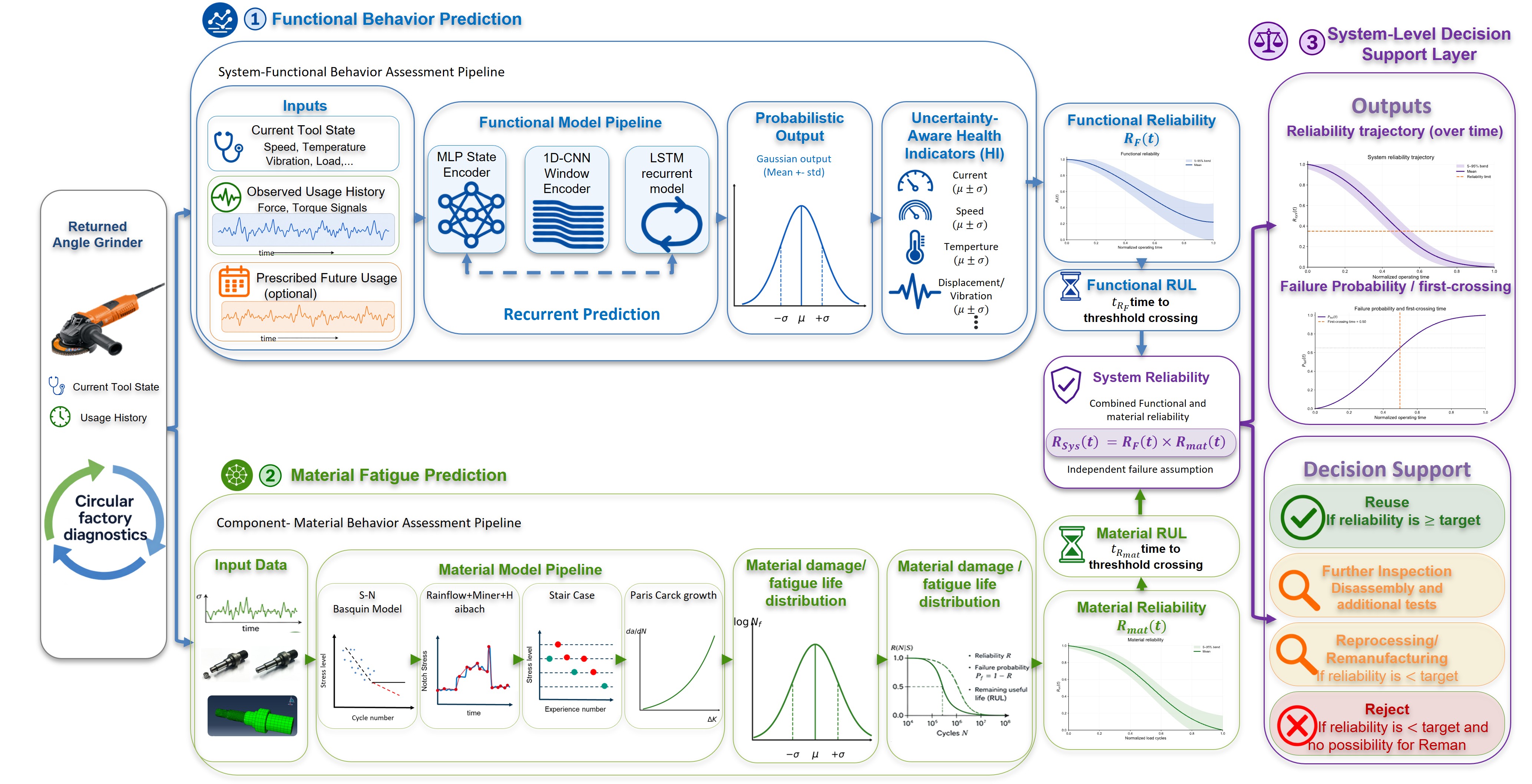}
    \label{fig:graphical_abstract}
\end{graphicalabstract}

% Keywords
% Each keyword is separated by \sep

\maketitle
% Main text
%%%%%%%%%%%%%%%%%%%%%%%%%%%%%%%%%%%%%%%%%%%%%%%%%%%%%%%%%%%%%%%%%%%%%%%%%%%%%%%%%
%%%%%%%%%%%%%%%%%%%%%%%%%%%%%%%%%%%%%%%%%%%%%%%%%%%%%%%%%%%%%%%%%%%%%%%%%%%%%%%%%
\section{Introduction}\label{}
The transition toward a circular economy requires manufacturing systems that retain product and material value across multiple life cycles rather than relying primarily on end-of-life recovery \citep{Kirchherr2017, EllenMacArthur2013, Geissdoerfer2017}. In this context, the Circular Factory (CF) envisions used products being transformed into new product generations through recombination, reconditioning, and targeted reuse of degraded subsystems \citep{Lanza2024}. Such reuse decisions must be instance-specific because returned products do not re-enter the factory in a uniform condition. Even nominally identical tools may differ in degradation history, operational loading, maintenance quality, and previous use conditions \citep{Guide2003, Grauberger2024}. Their future performance and failure risk are therefore heterogeneous, which challenges conventional reliability approaches based on homogeneous populations and fixed forward-life-cycle assumptions \citep{OConnor2012}.

For returned products, reliability cannot be assessed only in terms of complete structural failure. It must also be understood as continued adherence of function fulfillment over a subsequent service interval, consistent with the definition of reliability as the ability of a system to perform its intended function under specified conditions for a given period \citep{IEEE1991}. As unacceptable functional behavior may occur before catastrophic failure becomes visible, increased vibration, torque loss, thermal anomalies, or speed deviations may already make a returned product unsuitable for redeployment \citep{LeitenbergerCIRP2025}. The relevant question is therefore not only whether a tool still operates at inspection, but whether it can continue to fulfill its intended function under future usage.

Prognostics and health management (PHM) is relevant to this problem because condition and usage information can be used to estimate degradation and remaining useful life \citep{jardine2006review}. Health indicators (HIs) provide a compact representation of degradation progression and support future-state estimation \citep{Zhou2022}. However, existing prognostic approaches remain limited when operating conditions vary, usage histories are uncertain, and the physical meaning of the predicted degradation state is weak \citep{lei2018machinery, ariaschao2022fusing}. Moreover, system-level functional prediction alone is insufficient for circular redeployment, because reusable components may also accumulate material fatigue even while the product remains functional \citep{geist2024remanufactured}. A robust CF assessment must therefore combine future function fulfillment at the system level with material degradation at the component level.

In this paper, this assessment is investigated in a case study on a CF for angle grinders. This power tool serves as a representative CF use case because its reuse decision depends on both system-level behavior and component-level structural capability. A returned angle grinder may still operate, but its future suitability depends on how thermal, electrical, rotational, and geometric clearances evolve under continued use, and whether critical components, such as the output shaft, retain sufficient fatigue capacity. The case study, therefore, combines two complementary views: functional behavior prediction from the current state and recent force--torque usage history, and material-fatigue assessment of the output shaft from fatigue-relevant loading information. %\hl{The framework is positioned at the inspection and service-decision level of the Circular Factory. Its functional and material outputs support, but do not autonomously determine, the subsequent routing of returned products.} 
By linking these perspectives through shared condition and loading information, the framework assesses both the functional response under use and the structural capability required to sustain it.

In view of these challenges, this paper addresses the following research question:
\textbf{\emph{How can the current state, usage history, and future usage scenario of a returned product be represented in a unified functional–material reliability space to support instance-specific redeployment decisions under uncertainty?}}
%\textbf{\emph{How can the future functional behavior state and material-fatigue state of a returned product (angle grinder) be jointly assessed from current condition and usage history to support redeployment decisions in the CF?}}

To address this question, this paper introduces a framework for jointly evaluating future function fulfillment through uncertainty-aware system-level HI prediction and material-fatigue assessment of a critical component. The remainder of this paper is structured as follows. Section 2 reviews related work, Section 3 summarizes the contribution, Section 4 presents the methodology, Section 5 describes the case study and experimental setup, Sections 6 and 7 present and discuss the results, and Section 8 concludes the paper.
%%%%%%%%%%%%%%%%%%%%%%%%%%%%%%%%%%%%%%%%%%%%%%%%%%%%%%%%%%%%%%%%%%%%%%%%%%%%%%%%%
%%%%%%%%%%%%%%%%%%%%%%%%%%%%%%%%%%%%%%%%%%%%%%%%%%%%%%%%%%%%%%%%%%%%%%%%%%%%%%%%%
\section{Related Work}\label{}
Relevant prior work for this study can be organized into two connected strands. The first addresses how future function fulfillment can be inferred from operational data through degradation assessment, health indicators (HIs), and remaining-useful-life estimation. The second addresses how the remaining structural integrity of critical components can be evaluated from material degradation models under cyclic loading. Both perspectives are required for redeployment decisions in the Circular Factory (CF), because a returned product must remain suitable at both the system level and the component level.

\subsection{Functional Degradation Assessment from Operational Data}
Within prognostics and health management (PHM), condition and usage data are used to characterize degradation and estimate remaining useful life \citep{jardine2006review, Soualhi2022}. HIs are central in this process because they provide a compact representation of degradation progression and connect raw measurements to future-state prediction \citep{Zhou2025}. Their quality strongly affects the reliability of downstream prognostic estimates \citep{Kumar2024}. For returned products in the CF, this is especially relevant because redeployment depends not only on current operability, but also on how functional behavior is expected to evolve under further use.

Existing HI methods differ in how they represent degradation. Signal-based approaches derive degradation-sensitive features from time-domain, frequency-domain, or time-frequency representations of measured signals such as vibration \citep{jardine2006review, lei2018machinery}. These methods are often interpretable, but their diagnostic value can deteriorate when speed, load, or usage patterns vary. Model-based approaches infer a latent health state from physics-based, stochastic, or observer-based formulations, such as Wiener-process models or Kalman-type filters \citep{cubillo2016review, si2011remaining}. They provide physical interpretability and uncertainty-aware life estimates, but require substantial prior knowledge and often need re-identification when transferred to new systems \citep{an2015practical}. Data-driven approaches, including feature-fusion models, autoencoders, and recurrent sequence-learning architectures, learn degradation representations directly from historical data \citep{gonzalez2022autoencoder, wang2020jms, guo2017recurrent, yan2022deep}. They are attractive for systems whose degradation mechanisms are difficult to model analytically, but their reliability depends strongly on the representativeness of the training data \citep{lei2018machinery, zhan2023novel}.

Recent hybrid and physics-informed methods attempt to improve robustness by embedding prior knowledge, monotonicity assumptions, or damage-accumulation constraints into learned representations \citep{deng2023piml, ariaschao2022fusing, yucesan2020pinn}. Such constraints can reduce spurious correlations, but they can also bias the learned representation if the assumed physics do not match the dominant failure mechanism \citep{bajarunas2023unsupervised}. Overall, PHM provides a strong basis for functional degradation assessment, but three limitations remain important for CF applications: many methods are validated under fixed or narrowly defined operating conditions, many are developed on component-level benchmark data rather than full-system behavior, and population-trained models do not directly support instance-specific returned products with heterogeneous histories and future usage scenarios \citep{johnson2010emerging, Wang2018, peng2023multi, wu2024autoencoder, ansari2025estimating}.

\subsection{Material Degradation and Fatigue-Based Lifetime Assessment}
Material-side lifetime assessment is commonly based on two established approaches: stress-life fatigue models and fracture-mechanics-based crack-growth models. Classical fatigue assessment uses stress-life relations, such as Basquin-type S--N curves, together with rainflow cycle counting and Palmgren--Miner damage accumulation. For variable-amplitude loading, Haibach-type extensions are commonly used to account for cycles below the conventional endurance limit \citep{haibachBetriebsfestigkeitVerfahrenUnd2006, pyttelFKMRichtlinieBruchmechanischerFestigkeitsnachweis2007}. This workflow is efficient and widely used in engineering practice, but it provides limited physical resolution of the underlying damage mechanism.

Fracture-mechanics approaches, particularly Paris-law-based crack-growth models, provide a more explicit description of crack propagation once an initial crack is assumed to exist \citep{Paris1963, schijveFatigueStructuresMaterials2010}. These models are useful for structural integrity assessment, but they require assumptions about initial crack size, critical crack length, geometry factors, and material parameters. They also describe the propagation phase more directly than the early initiation-dominated phase. In practice, fatigue assessment often combines stress-life, damage-accumulation, and crack-growth models, but the transition between initiation, propagation, and total life remains methodologically non-trivial.

For remanufacturing and circular manufacturing, this material perspective is essential because components intended for reuse can accumulate fatigue damage over multiple life cycles even when system-level function remains acceptable \citep{geist2024remanufactured}. However, material lifetime models are usually applied at the component level and are rarely linked to system-level functional prognosis using the same operational history. This separation limits their usefulness for CF redeployment decisions, where the same returned product must be assessed with respect to both future function fulfillment and remaining structural capability.

\subsection{Research Gap}
Taken together, existing work provides strong foundations for functional degradation prediction and material fatigue assessment, but these perspectives are still mostly treated separately. HI-based prognostic methods support future-state prediction from operational data, yet they often struggle with variable usage histories, full-system interactions, and instance-specific returned products. Material fatigue models provide physically grounded structural assessment, but they are typically component-level models and are not consistently integrated with system-level functional behavior prediction. What remains missing is a framework that uses shared condition and usage information to estimate both future functional behavior and material-fatigue progression under uncertainty. This gap motivates the present work, which jointly links recent operational history, current system state, functional reliability, and component-level material assessment for instance-specific redeployment decisions in the CF.
%%%%%%%%%%%%%%%%%%%%%%%%%%%%%%%%%%%%%%%%%%%%%%%%%%%%%%%%%%%%%%%%%%%%%%%%%%%%%%%%%
%%%%%%%%%%%%%%%%%%%%%%%%%%%%%%%%%%%%%%%%%%%%%%%%%%%%%%%%%%%%%%%%%%%%%%%%%%%%%%%%%
\section{Contribution}\label{}
In contrast to conventional prognostic formulations that predict a single health indicator or remaining useful life, the proposed framework first establishes a unified functional–material assessment space for the returned product. This space combines the current system state, recent usage history, and material degradation state into a common representation. Within this space, redeployment is formulated as an admissibility problem in which the assessed functional and material requirements must be satisfied under a prescribed future usage scenario. The LSTM-based functional model and the fatigue-based material model, therefore, serve as complementary estimators of the position and evolution of the returned product within this reliability space.

Building on this gap, the paper contributes an uncertainty-aware framework for assessing returned products from both system-function and component-material perspectives at the example of an angle grinder. The main contributions are:

\begin{enumerate}
\item Future functional behavior is formulated as a conditional sequence-learning problem that combines the current system state with recent usage history (force--torque at the angle grinder).
\item The model supports scenario-conditioned inference by prescribing future usage windows, enabling evaluation of functional behavior and reliability under explicit operating conditions.
\item An uncertainty-aware LSTM architecture with Gaussian negative log-likelihood training is used to predict both the expected functional response and its predictive uncertainty.
\item Component-level material fatigue is integrated through stress reconstruction, fatigue damage assessment, and crack-growth-based lifetime evaluation (at the output spindle shaft of the angle grinder).
\item The functional and material branches are combined in a streaming reliability procedure that updates functional and material reliability and a system reliability indicator for instance-specific redeployment decisions.
\end{enumerate}
Together, these contributions provide the first integrated assessment framework that links measured operational loading to both future functional fulfillment and remaining structural fatigue life within a single streaming workflow, thereby supporting instance-specific redeployment decisions in the circular factory.
%%%%%%%%%%%%%%%%%%%%%%%%%%%%%%%%%%%%%%%%%%%%%%%%%%%%%%%%%%%%%%%%%%%%%%%%%%%%%%%%%
%%%%%%%%%%%%%%%%%%%%%%%%%%%%%%%%%%%%%%%%%%%%%%%%%%%%%%%%%%%%%%%%%%%%%%%%%%%%%%%%%
\section{Methodology}\label{}
The methodology translates the circular-factory redeployment problem into an operational prediction framework. The framework links the current system state and recent usage history to future functional behavior, while also connecting the same loading history to material-fatigue assessment of the component. The component-to-system organization follows the architectural perspective of \citet{Afifi2026RAMS}, which relates component failure mechanisms to functional behavior and system reliability. The following subsections describe the technical methodology and model building for the functional prediction pipeline and the material-fatigue pipeline, and their integration into a streaming reliability analysis.
%Building on the angle-grinder use case, the methodology translates the circular-factory redeployment problem into an operational prediction framework. The framework links the current system state and recent usage history to future functional behavior, while also connecting the same loading history to material-fatigue assessment of the output shaft. The following subsections describe the functional prediction pipeline, the material-fatigue pipeline, and their integration in a streaming reliability analysis.

%%%%%%%%%%%%%%%%%%%%%%%%%%%%%%%%%%%%%%%%%%%%%%%%%%%%%%%%%%%%%%%%%%%%%%%%%%%%%%%%%%%%%%%%%%%
\subsection{System Functional Behavior Prediction Pipeline}
\subsubsection{State, Usage History, and Health Indicator Variables}
The functional model uses two input types: the current system state and a short history of recent loading. The measured and derived quantities are therefore organized into three groups. \textbf{State variables} describe the thermo-mechanical and geometric condition at the start of the prediction. \textbf{Usage-history variables} describe the recent operational excitation over preceding time windows, including both load magnitude and temporal pattern. \textbf{Health-indicator variables} describe degradation-relevant quantities used for prediction and reliability interpretation. In the angle-grinder use case, pinion and spindle clearances serve as health indicators because increasing mechanical play reflects wear progression in bearing and gear-support interfaces.
%%%%%%%%%%%%%%%%%%%%%%%%%%%%%%%%%%%%%%%%%%%%%%%%%%%%%%%%%%%%%%%%%%%%%%%%%%%%%%%%%%%%%%%%%%%
\subsubsection{Prediction Setup}
At prediction time \(t\), the model receives a state vector \(\mathbf{s}_t \in \mathbb{R}^{d_s}\) and a sequence of recent usage windows,
\[
\mathbf{U}_{t-L+1:t} = \left(\mathbf{U}_{t-L+1}, \ldots, \mathbf{U}_{t}\right),
\]
where each \(\mathbf{U}_k \in \mathbb{R}^{P \times d_u}\) contains denoised and resampled force- and torque-related signals over a fixed time interval. The prediction target is the functional behavior vector \(\mathbf{y}_t \in \mathbb{R}^{d_y}\), which contains thermal, electrical, rotational, and geometry-related condition variables.

The learning problem is formulated as
\[
    p_{\theta}\left(\mathbf{y}_t \mid \mathbf{s}_t, \mathbf{U}_{t-L+1:t}\right)
\]
where \(\theta\) denotes the trainable model parameters. This captures the main assumption that future functional behavior depends jointly on the current tool state and recent usage history.

The loading-history representation preserves load magnitudes, transient events, and temporal patterns. Together with the observed state, these inputs support an end-to-end prediction of functional behavior without prescribing component-specific degradation laws. Combined degradation-related effects can be represented when they are captured by the selected variables and represented in the training data, without identifying the contribution of each underlying mechanism.

The network predicts increments rather than absolute target values. The absolute trajectory is reconstructed by adding the predicted changes to the measured initial value. This keeps the forecast anchored to the known starting condition and lets the model focus on local evolution, which is typically smoother than the absolute signal level.

For scenario-conditioned inference, additional future usage windows
\[
\widetilde{\mathbf{U}}_{t+1:t+H} = \left(\widetilde{\mathbf{U}}_{t+1}, \ldots, \widetilde{\mathbf{U}}_{t+H}\right)
\]
can be prescribed. The model then estimates the future response under this operating condition as
\[
    p_{\theta}\left(\mathbf{Y}_{t+1:t+H} \mid \mathbf{s}_t, \mathbf{U}^{\mathrm{obs}}_{t-L+1:t}, \widetilde{\mathbf{U}}_{t+1:t+H}\right)
\]
This allows functional behavior and reliability to be evaluated under known or planned future loading profiles rather than only under historically observed usage.
%%%%%%%%%%%%%%%%%%%%%%%%%%%%%%%%%%%%%%%%%%%%%%%%%%%%%%%%%%%%%%%%%%%%%%%%%%%%%%%%%%%%%%%%%%%%%%%%%%%%%
\subsubsection{Data Representation}
\begin{figure}
    \centering
    \includegraphics[width=0.6\linewidth]{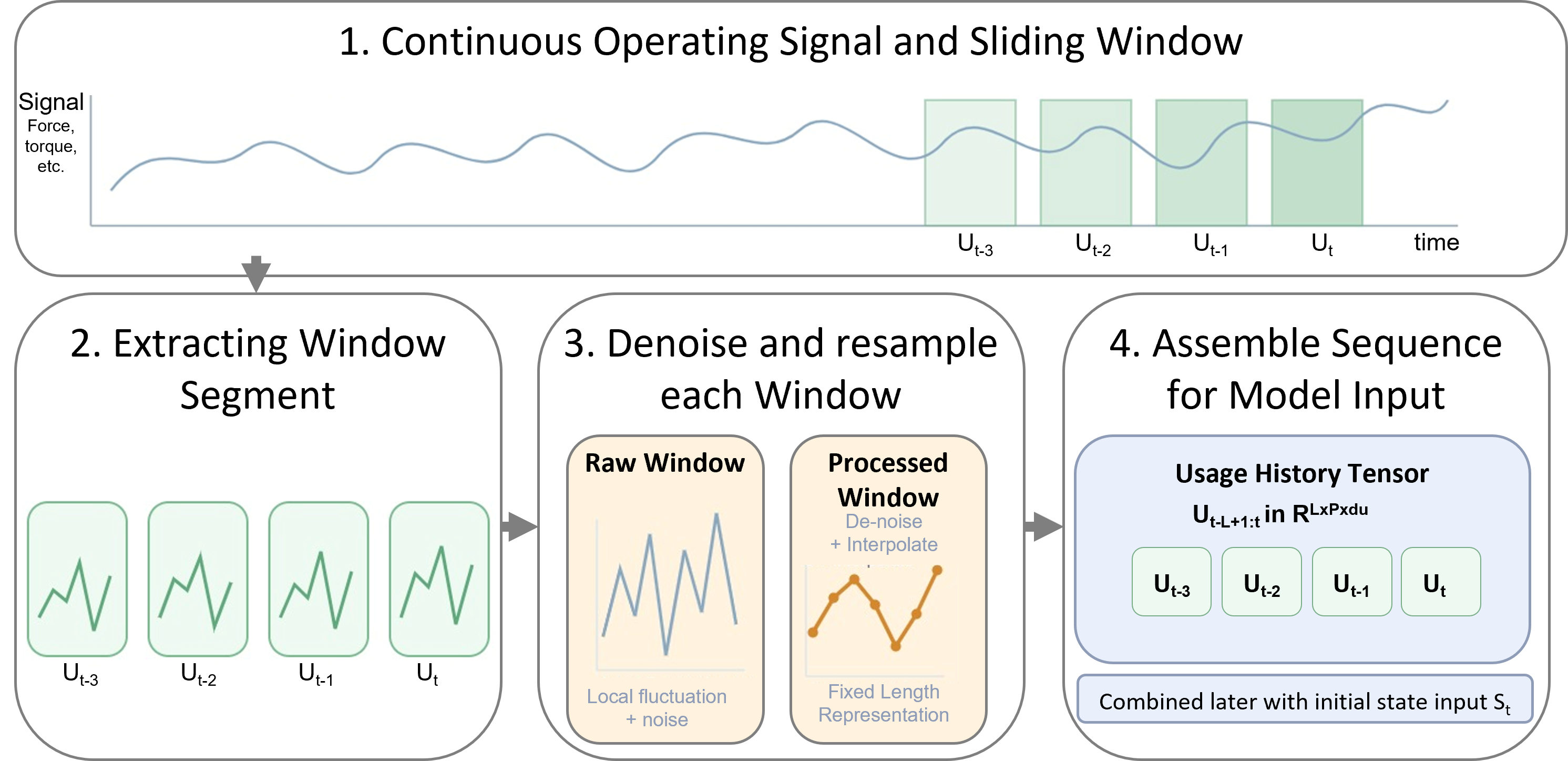}
    \caption{Data Preprocessing and Segmentation Framework}
    \vspace{-5mm}
    \label{fig:data_preprocessing}
\end{figure}

The data representation follows the conditional prediction setup. Each sample consists of one state vector, a sequence of recent usage-history windows, and the corresponding target value or target sequence. Continuous force- and torque-related measurements are segmented into sliding windows so that each window preserves local temporal patterns, while consecutive windows preserve recent usage evolution. Anchor points are placed at equal time intervals by sliding the window scheme over each recorded run. For each valid anchor point, the state vector is paired with the most recent usage windows and the corresponding target. The same representation is used during training and inference, as illustrated in Fig.~\ref{fig:data_preprocessing}.
%%%%%%%%%%%%%%%%%%%%%%%%%%%%%%%%%%%%%%%%%%%%%%%%%%%%%%%%%%%%%%%%%%%%%%%%%%%%%%%%%%%%%%%%%%%%%%%%%%%%%
%%%%%%%%%%%%%%%%%%%%%%%%%%%%%%%%%%%%%%%%%%%%%%%%%%%%%%%%%%%%%%%%%%%%%%%%%%%%%%%%%%%%%%%%%%%%%%%%%%%%%
\subsubsection{Model Architecture}
\begin{figure}
    \centering
    \includegraphics[width=0.7\linewidth, trim=0.75cm 4.8cm 0.65cm 14.2cm, clip]{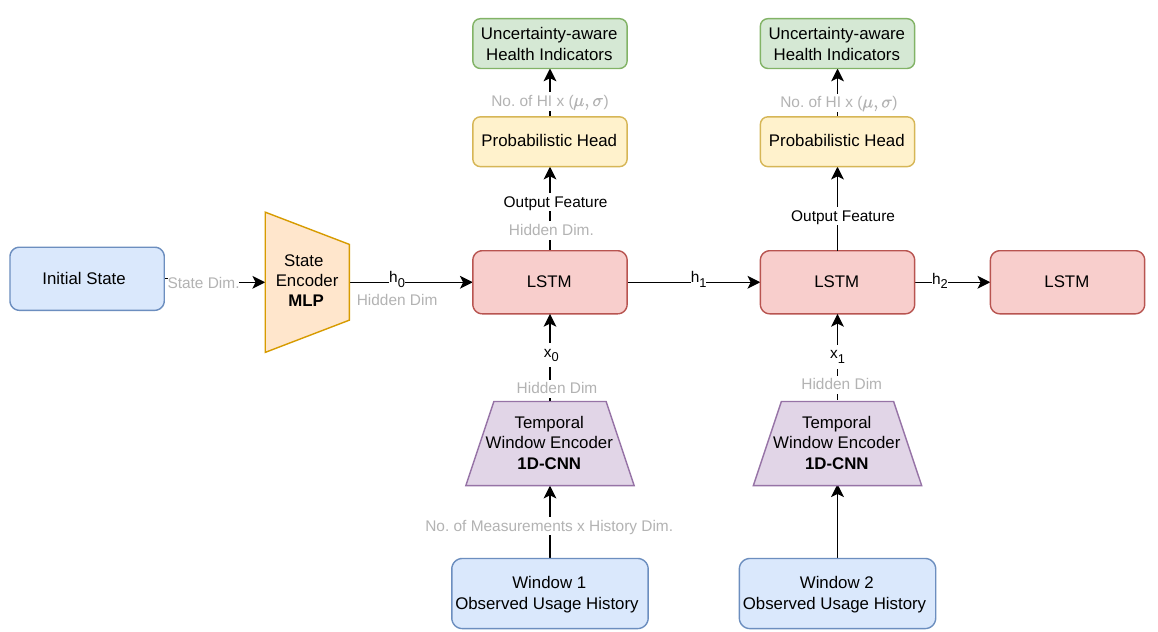}
    \caption{Architecture of the uncertainty-aware functional behavior prediction model. The current state vector and usage history windows are encoded in separate branches, fused, processed by an LSTM recurrent backbone, and mapped to predictive means and variances for the functional output variables.}
    \label{fig:model_arch}
\end{figure}

The model consists of a usage-history encoder, a state encoder, a recurrent sequence model, and an uncertainty-aware output head, as shown in Fig.~\ref{fig:model_arch}. For each usage window, 1D convolutional layers extract local temporal patterns from the force--torque history, such as peaks, transients, and short load variations. In parallel, a multi-layer perceptron maps the state vector to a compact representation of the current tool condition. The two representations are then fused and processed by an LSTM recurrent backbone. The LSTM is used because the present task involves short sequences of overlapping usage windows whose local structure is already summarized by the convolutional encoder. Under these conditions, the recurrent block mainly needs to preserve recent load context and integrate it with the current state, which favors a simple and stable LSTM over more elaborate recurrent variants. The final head outputs a predictive mean and variance for each target variable, enabling both point prediction and threshold-based reliability assessment.
%%%%%%%%%%%%%%%%%%%%%%%%%%%%%%%%%%%%%%%%%%%%%%%%%%%%%%%%%%%%%%%%%%%%%%%%%%%%%
%%%%%%%%%%%%%%%%%%%%%%%%%%%%%%%%%%%%%%%%%%%%%%%%%%%%%%%%%%%%%%%%%%%%%%%%%%%%%
\subsubsection{Uncertainty-Aware Loss Function}
The model is trained with a weighted Gaussian negative log-likelihood loss because it predicts both an expected value and an uncertainty estimate. For sample \(n\), sequence position \(\ell\), and output variable \(m\), let \(y_{n\ell m}\) denote the target, and let \(\mu_{n\ell m}\) and \(\sigma^2_{n\ell m}\) denote the predicted mean and variance. The loss is
\begin{equation}
    \mathcal{L}(\theta) = \frac{1}{N} \sum_{n,\ell,m} w_m \left[ \log \sigma^2_{n\ell m} + \frac{\left(y_{n\ell m} - \mu_{n\ell m}\right)^2}{\sigma^2_{n\ell m}} \right],
\end{equation}
where \(N\) is the number of training samples and \(w_m\) is the weight assigned to output variable \(m\). The squared-error term penalizes inaccurate predictions, while the logarithmic term discourages excessively large variances. This is important for reliability assessment because both the predicted mean and uncertainty determine the probability of functional unfulfillment.
%%%%%%%%%%%%%%%%%%%%%%%%%%%%%%%%%%%%%%%%%%%%%%%%%%%%%%%%%%%%%%%%%%%%%%%%%%%%%%
\subsubsection{Training--Testing Framework}
\textbf{Training Algorithm}
Algorithm~\ref{alg:training} summarizes training. The model encodes the state and usage history, predicts target increments and variances, reconstructs absolute outputs from the observed initial value, and optimizes the weighted Gaussian negative log-likelihood.
\begin{algorithm}[h]
\caption{Training procedure of the framework}
\label{alg:training}
\begin{algorithmic}[1]
\Require Mini-batches \(\mathcal{B}=\{(\mathbf{s}^{(n)}_t,\mathbf{U}^{(n)}_{t-L+1:t},\mathbf{Y}^{(n)}_{t-L:t})\}_{n=1}^{N}\)
\State Initialize trainable parameters \(\theta\)
\For{each training epoch}
    \For{each mini-batch \(\mathcal{B}\)}
        \State encode \(\mathbf{s}^{(n)}_t\) and \(\mathbf{U}^{(n)}_{t-L+1:t}\), fuse both representations, and process them with the recurrent backbone
        \State predict mean increments \(\widehat{\Delta \mathbf{y}}^{(n)}_{t-L+\ell}\) and predictive variances \(\widehat{\boldsymbol{\sigma}}^{2,(n)}_{t-L+\ell}\) for \(\ell=1,\dots,L\)
        \State reconstruct absolute predictions from the observed initial state using \(\widehat{\mathbf{y}}^{(n)}_{t-L+\ell}=\mathbf{y}^{(n)}_{t-L}+\sum_{j=1}^{\ell}\widehat{\Delta \mathbf{y}}^{(n)}_{t-L+j}\) 
        \State evaluate the weighted Gaussian negative log-likelihood \(\mathcal{L}(\theta)\)
        \State update \(\theta\) with AdamW
    \EndFor
\EndFor
\State keep the model \(p_{\theta}\) with the best validation performance
\end{algorithmic}
\end{algorithm}

%%%%%%%%%%%%%%%%%%%%%%%%%%%%%%%%%%%%%%%%%%%%%%%%%%%%%%%%%%%%%%%%%%%%%%%%%%%%%%
\textbf{Evaluation Algorithm}
Evaluation is performed on held-out files only, preserving file-level separation between training and testing. As shown in Algorithm~\ref{alg:evaluation}, predictions, variances, and observations are accumulated over the test set and then summarized for each output variable.
%%%%%%%%%%%%%%%%%%%%%%%%%%%%%%%%%%%%%%%%%%%%%%%%%%%%%%%%%%%%%%%%%%%%%%%%%%%%%%

\begin{algorithm}[h]
\caption{Held-out evaluation on the test set}
\label{alg:evaluation}
\begin{algorithmic}[1]
\Require trained model \(p_{\theta}\), held-out test set \(\mathcal{D}_{\mathrm{test}}\)
\State initialize storage \(\mathcal{S}_m\) for each output variable \(m=1,\dots,d_y\)
\For{each test sample \(n \in \mathcal{D}_{\mathrm{test}}\)}
    \State encode \(\mathbf{s}^{(n)}_t\) and \(\mathbf{U}^{(n)}_{t-L+1:t}\), fuse both representations, and process them with the recurrent backbone
    \State predict mean increments \(\widehat{\Delta \mathbf{y}}^{(n)}_{t-L+\ell}\) and predictive variances \(\widehat{\boldsymbol{\sigma}}^{2,(n)}_{t-L+\ell}\) for \(\ell=1,\dots,L\)
        \State reconstruct absolute predictions from the observed initial state using \(\widehat{\mathbf{y}}^{(n)}_{t-L+\ell}=\mathbf{y}^{(n)}_{t-L}+\sum_{j=1}^{\ell}\widehat{\Delta \mathbf{y}}^{(n)}_{t-L+j}\) 
    \For{each output variable \(m\)}
        \State store the reconstructed predictions, predictive variances, and observed values of indicator \(m\) in \(\mathcal{S}_m\)
    \EndFor
\EndFor
\For{each output variable \(m\)}
    \State calculate the desired evaluation metrics from the stored predictions, predictive variances, and targets in \(\mathcal{S}_m\)
    \State average each metric over all held-out samples and supervised sequence positions
\EndFor
\State report aggregated regression and reliability results on \(\mathcal{D}_{\mathrm{test}}\)
\end{algorithmic}
\end{algorithm}
%%%%%%%%%%%%%%%%%%%%%%%%%%%%%%%%%%%%%%%%%%%%%%%%%%%%%%%%%%%%%%%%%%%%%%%%%%%%%%

\subsection{Component--Material Behavior Prediction Pipeline}
Material-side reuse assessment evaluates the remaining fatigue capability of the output shaft from the same operational load history used for functional prediction. The pipeline combines three steps: finite element analysis for local stress reconstruction, a stress--life fatigue characterization of the shaft under rotating bending, and a fracture-mechanics-based crack-growth model. Together, these steps translate measured force--torque windows into cumulative material damage and a corresponding material reliability estimate.
\begin{figure}
    \centering    
    \includegraphics[width=0.95\linewidth]{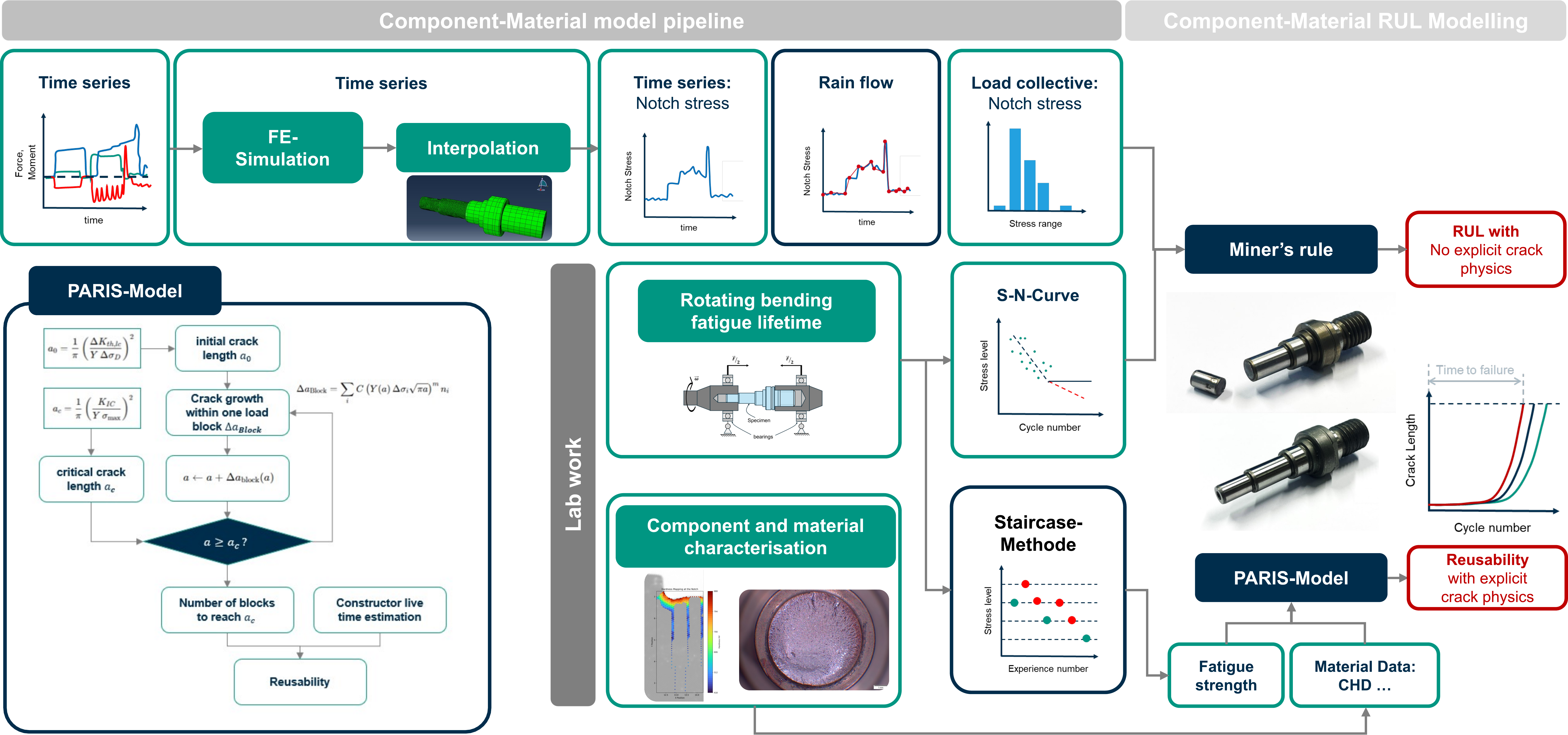}
    \caption{Component--Material Behavior Prediction Pipeline}
    \label{fig:caseStudy}
\end{figure}
\subsubsection{Finite Element Analysis and Interpolation-Based Stress Reconstruction}
Finite element analysis was conducted using Abaqus to determine the local stress distribution in the output shaft under rotating bending conditions. A three-dimensional model of the shaft geometry was implemented based on the technical drawing. The material behavior was assumed to be linear elastic, characterized by Young’s modulus and Poisson’s ratio corresponding to the investigated steel grade. Boundary conditions were defined to replicate the rotating bending configuration by applying a bending moment at the critical cross-section while constraining the shaft ends accordingly. The mesh was locally refined in the notch region to accurately capture stress gradients. The resulting stress field was evaluated in terms of the local notch stress, which served as input for the fatigue and crack propagation analyses.
In addition to the rotating bending simulation, a separate numerical study was carried out to determine the stress response of the shaft under service-related loading conditions. For this purpose, a simulation matrix was generated based on six independent load components. The parameter space was sampled using Latin hypercube sampling in order to achieve an efficient and well-distributed coverage of the admissible load range. Two additional bending moments were derived directly from the corresponding force components using the associated lever arms.
For each load case, the local maximum stress at the critical notch location was extracted from the finite element results. Based on these simulation data, an interpolation model was established to map the external load components to the local stress response. This interpolation approach was subsequently used to transform measured or prescribed force time histories into a corresponding stress-time history at the notch root.

\subsubsection{Fatigue-life assessment framework}
The fatigue-life assessment was performed using a combined stress--life and fracture-mechanics-based approach. The S--N behavior in the initiation-dominated regime was described by a Basquin relationship fitted to the rotating bending experimental data. For this purpose, the notch stress amplitudes obtained from the fatigue tests were correlated with the corresponding numbers of cycles to failure in logarithmic representation. In addition to the original S--N data, a shifted S--N curve was derived by subtracting the Paris-law crack propagation life from the experimentally observed total fatigue life. This separation isolates the initiation-dominated portion of the lifetime from the propagation-dominated portion, allowing the Basquin fit to characterize crack initiation independently and thereby avoiding overestimation of the initiation life when the propagation contribution is non-negligible. Basquin fits were determined for both the original and the shifted S--N data. For variable-amplitude loading, the service-related stress histories at the critical notch location were evaluated using rainflow cycle counting, followed by linear damage accumulation according to the Palmgren--Miner rule. To account for load cycles below the conventional endurance limit, the S--N curve was extended using the Haibach modification. Based on this procedure, the cumulative damage and the corresponding number of repetitions of the load spectrum up to failure were determined.

The crack propagation contribution was assessed separately using a fracture-mechanics-based model. The propagation life calculated from the Paris law was assigned to each stress level and subtracted from the total fatigue life to obtain the initiation-related lifetime. The derivation of the crack-growth parameters is given in the following subsection.

\subsubsection{Crack initiation--propagation transition and modeling}
The Paris-law propagation life subtracted from the total fatigue life in the framework above is derived from the following crack-growth model. The transition from crack initiation to crack propagation was defined based on a threshold in terms of the stress intensity factor range. The initial crack size $a_0$ was determined from the long-crack threshold $\Delta K_{\mathrm{th,lc}}$ using an inverse formulation of the stress intensity factor. To avoid applying long-crack concepts in the physically short-crack regime, a lower bound for $a_0$ was imposed. This lower bound was chosen on the order of ten times a characteristic microstructural length scale, here approximated by the prior-austenite grain size of the tempered martensitic microstructure. This assumption represents an engineering approximation to ensure that the crack growth analysis starts within the long-crack regime.

In addition, an upper limit for the initial crack size was introduced to restrict the defect size to physically realistic values. The effective initial crack length used in the crack growth calculation was therefore defined as
\begin{equation}
a_0 = \min \left( a_{0,\mathrm{calc}}, \; a_{0,\mathrm{limit}} \right)
\end{equation}

The critical crack length $a_c$ was obtained from the fracture condition $K_I = K_{\mathrm{IC}}$ under the assumption of linear-elastic fracture mechanics. To account for material and geometric constraints, the calculated critical crack length was limited by the effective case-hardened layer thickness. This ensures that crack propagation is restricted to the mechanically relevant hardened zone. The effective critical crack length was therefore defined as
\begin{equation}
a_c = \min \left( a_{c,\mathrm{calc}}, \; a_{\mathrm{case}} \right)
\end{equation}

Fatigue crack growth between $a_0$ and $a_c$ was modeled using the Paris law, expressed as a function of the stress-intensity-factor range. The stress intensity factor range $\Delta K$ was calculated as
\begin{equation}
\Delta K = Y \, \Delta \sigma \, \sqrt{\pi a}
\end{equation}
assuming a constant geometry factor $Y$. The stress range $\Delta \sigma$ was derived from the local stress amplitude obtained via finite element simulations at the notch root under fully reversed loading conditions ($R = -1$).

The crack propagation life $N_{\mathrm{prop}}$ was determined by numerical integration of the Paris equation over the crack length interval from $a_0$ to $a_c$:
\begin{equation}
N_{\mathrm{prop}} = \int_{a_0}^{a_c} \frac{1}{C \left( \Delta K \right)^m} \, \mathrm{d}a
\end{equation}

A discretized crack-length approach was used, in which the crack-length domain was subdivided into a large number of increments, and the corresponding cycle increments were accumulated. This procedure enables a stable and reproducible evaluation of crack growth life across different stress levels. The material parameters used for crack propagation modeling are summarized in Table~\ref{tab:paris_parameters}. 

\begin{table}[t]
\vspace{-5mm}
\caption{Material parameters used for fatigue crack growth modeling  \citep{liNotchFatigueLife2023}.}
\vspace{-1.5mm}
\label{tab:paris_parameters}
\centering
\begin{tabular}{llll}
\toprule
Parameter & Symbol & Value & Unit \\
\midrule
Fracture toughness & $K_{\mathrm{IC}}$ & 131.56 & $\mathrm{MPa}\sqrt{\mathrm{m}}$ \\
Effective threshold SIF range & $\Delta K_{\mathrm{th,eff}}$ & 2.5 & $\mathrm{MPa}\sqrt{\mathrm{m}}$ \\
Paris coefficient & $C$ & $3.5882 \times 10^{-11}$ & $\mathrm{m\,cycle^{-1}}(\mathrm{MPa}\sqrt{\mathrm{m}})^{-m}$ \\
Paris exponent & $m$ & 2.5022 & -- \\
\bottomrule
\end{tabular}
\end{table}
\begin{comment}
\begin{table}[width=\linewidth,pos=t]
\vspace{-5mm}
\caption{Material parameters used for fatigue crack growth modeling.}
\vspace{-1.5mm}
\label{tab:paris_parameters}
\centering
\begin{tabular}{lllll}
\toprule
Parameter & Symbol & Value & Unit \\
\midrule
Fracture toughness & $K_{\mathrm{IC}}$ & 131.56 & MPa\,\sqrt{\mathrm{m}}  \\
effective threshold SIF range & $\Delta K_{\mathrm{th,eff}}$ & 2.5 & MPa\,\sqrt{\mathrm{m}} \\
Paris coefficient  & $C$ & $3.5882 \times 10^{-11}$ & $\mathrm{m\,cycle^{-1}}\,(\mathrm{MPa}\sqrt{\mathrm{m}})^{-m}$  \\
Paris exponent     & $m$ & 2.5022 & --  \\
\bottomrule
\end{tabular}
\end{table}
\end{comment}
%%%%%%%%%%%%%%%%%%%%%%%%%%%%%%%%%%%%%%%%%%%%%%%%%%%%%%%%%%%%%%%%%%%%%%%%%%%%%%%%%%%%%%%%%%%%
\subsection{Consolidation of Functional-Material Reliability Algorithm}
The final assessment step consolidates the full methodology into a single streaming reliability analysis, as demonstrated in Algorithm \ref{alg:combined_reliability}. A trained functional model and a material model are assumed to be available before deployment. For each new window in the stream data, the accumulated state and the incoming usage window are evaluated jointly. On the functional side, the procedure follows the same logic as in Algorithms~\ref{alg:training} and~\ref{alg:evaluation}: the state and usage window are encoded, fused, and processed with the recurrent backbone to obtain a hidden state, from which the predictive mean and variance are inferred and the absolute health-indicator values are reconstructed from the observed initial state. On the material side, the same window is mapped to incremental material damage and added to the cumulative damage state. Because streamed windows may not be consecutive in cycle time, any gap between adjacent windows is accounted for by propagating the most recent incremental damage estimate over the missing cycles. In this way, the conditional prediction setup, the sliding-window representation, the uncertainty-aware recurrent model, and the material-fatigue branch are combined into a single step that provides updated values for the functional reliability summary \(R_{\mathrm{func}}\), material reliability \(R_{\mathrm{mat}}\), system reliability indicator \(R_{\mathrm{sys}}\), cumulative damage quantiles, and the governing branch after each processed window. The governing branch identifies which reliability dimension is binding at each inspection point, functional or material, providing a direct indication of whether the redeployment decision is limited by predicted behavioral degradation or by accumulated structural fatigue.
%%%%%%%%%%%%%%%%%%%%%%%%%%%%%%%%%%%%%%%%%%%%%%%%%%%%%%%%%%%%%%%%%%%%%%%%%%%%%%%%%%%%
Here, \(R_{\mathrm{func}}\) summarizes the least favorable individual functional probability, and \(R_{\mathrm{sys}}=\min(R_{\mathrm{func}},R_{\mathrm{mat}})\) serves as a system reliability indicator identifying the limiting assessed requirement. The minimum operation summarizes the modeled criteria without specifying their joint probability distribution or calculating complete-machine survival probability. No statistical independence assumption is introduced by this aggregation. The retained functional probabilities concern the selected prediction position, whereas material reliability concerns damage accumulated up to the inspection. Their minimum therefore provides an inspection-level assessment summary. The crack-growth-based reuse estimates are evaluated alongside the Miner-damage replay and reported in Section~6.5.

\begin{algorithm}[t]
\caption{Consolidation of functional-material reliability analysis}
\label{alg:combined_reliability}
\begin{algorithmic}[1]
\Require trained functional model \(p_{\theta}\), calibrated material model \(\mathcal{M}_{\mathrm{mat}}\), stream data \(\mathcal{W}=\{(\mathbf{s}_i,\mathbf{U}_i,\mathbf{y}_{i,0},\mathcal{F}_i,c_i)\}_{i=1}^{N}\), functional thresholds \(\{\tau_m\}_{m=1}^{d_y}\), bootstrap count \(B\)
\State initialize cumulative material-damage samples \(\mathbf{d} \leftarrow \mathbf{0} \in \mathbb{R}^{B}\)
\State initialize storage \(\mathcal{S}\) for window-wise functional, material, and system summaries
\For{each streamed window \(i=1,\dots,N\)}
    \State encode \(\mathbf{s}_i\) and \(\mathbf{U}_i\), fuse both representations, and process them with the recurrent backbone to obtain hidden state \(\mathbf{h}_i\)
    \State predict mean increments \(\widehat{\Delta \mathbf{y}}_{i,\ell}\) and predictive variances \(\widehat{\boldsymbol{\sigma}}^{2}_{i,\ell}\) from \(\mathbf{h}_i\) for \(\ell=1,\dots,L\)
    \State reconstruct absolute predictions from the observed anchor state using \(\widehat{\mathbf{y}}_{i,\ell}=\mathbf{y}_{i,0}+\sum_{j=1}^{\ell}\widehat{\Delta \mathbf{y}}_{i,j}\)
    \State retain the last reconstructed prediction \(\widehat{\boldsymbol{\mu}}_i=\widehat{\mathbf{y}}_{i,L}\) and the corresponding predictive standard deviation \(\widehat{\boldsymbol{\sigma}}_i\)
    \For{each output variable \(m=1,\dots,d_y\)}
        \State compute functional failure probability \(q_{i,m}=1-\Phi\!\left(\frac{\tau_m-\widehat{\mu}_{i,m}}{\widehat{\sigma}_{i,m}}\right)\)
        \State set functional reliability \(R_{i,m}=1-q_{i,m}\)
    \EndFor
    \State summarize the functional branch as \(R_{\mathrm{func},i}=\min_{m} R_{i,m}\)
    \State evaluate the material model on \(\mathcal{F}_i\) and obtain window material damage samples \(\Delta \mathbf{d}_i\)
    \State update cumulative material damage \(\mathbf{d} \leftarrow \mathbf{d} + \Delta \mathbf{d}_i\)
    \State estimate material reliability as \(R_{\mathrm{mat},i}=\frac{1}{B}\sum_{b=1}^{B}\mathbb{I}[d_b < 1]\)
    \If{\(i < N\) and gap \(g_i = c_{i+1}-c_i-1 > 0\)}
        \State propagate missing-cycle damage: \(\mathbf{d} \leftarrow \mathbf{d} + g_i\,\Delta\mathbf{d}_i\)
    \EndIf
    \State compute the system reliability indicator \(R_{\mathrm{sys},i}=\min(R_{\mathrm{func},i},R_{\mathrm{mat},i})\)
    \State identify governing branch \(\mathit{branch}_i = \arg\min(R_{\mathrm{func},i},\,R_{\mathrm{mat},i})\)
    \State append \((c_i,\,R_{\mathrm{func},i},\,R_{\mathrm{mat},i},\,R_{\mathrm{sys},i},\,d^{(5)}_i,\,d^{(50)}_i,\,d^{(95)}_i,\,\mathit{branch}_i)\) to \(\mathcal{S}\)
\EndFor
\State report the streaming trajectories of \(R_{\mathrm{func}}\), \(R_{\mathrm{mat}}\), \(R_{\mathrm{sys}}\), cumulative damage quantiles, and the governing branch over the full sequence
\end{algorithmic}
\end{algorithm}
%%%%%%%%%%%%%%%%%%%%%%%%%%%%%%%%%%%%%%%%%%%%%%%%%%%%%%%%%%%%%%%
\section{Case Study: Angle Grinder}
The case study considered in this work is an angle grinder, selected as a representative returned product for instance-specific functional and material assessment in a circular-factory context. Functional behavior is investigated using a gear-stage test bench with an external drive motor and controlled force--torque loading. The corresponding material assessment evaluates output-shaft reliability, linked to system behavior through the shaft's role in transmitting torque, for which structural integrity is a necessary requirement. This case study provides the basis for linking functional behavior prediction with component-level material assessment, as illustrated in Fig.~\ref{fig:caseStudy}.

\textbf{Assessment assumptions.} Functional prediction is conditioned on the selected current-state and usage-history variables defined in Section~5.1.1. Validation covers the investigated repeating load cycle, while transfer to other operating profiles and returned-product states requires further validation. The Gaussian outputs describe conditional predictive uncertainty for the selected variables, and the empirical thresholds in Section~5.1.4 define study-specific functional assessment events.

Material-fatigue assessment concerns the output shaft. Stress reconstruction assumes linear-elastic behavior, and Miner accumulation neglects load-sequence effects. Paris-law estimates depend on the assumed initial defect, crack-growth boundaries, geometry factor, and literature-based material parameters described in Section~4.2. The Miner-damage replay starts from zero cumulative damage and approximates gaps using the most recent incremental damage estimate. Its damage estimates therefore describe accumulation from that baseline rather than unknown prior damage. Shared loading connects the assessments, while component-specific degradation interactions and joint failure-event dependencies are not explicitly quantified.

%%%%%%%%%%%%%%%%%%%%%%%%%%%%%%%%%%%%%%%%%%%%%%%%%%%%%%%%%%%%%%%%%%%%%%%%%%%%%%%%%%%%
\begin{figure}
    \centering
    \includegraphics[width=0.52\linewidth]{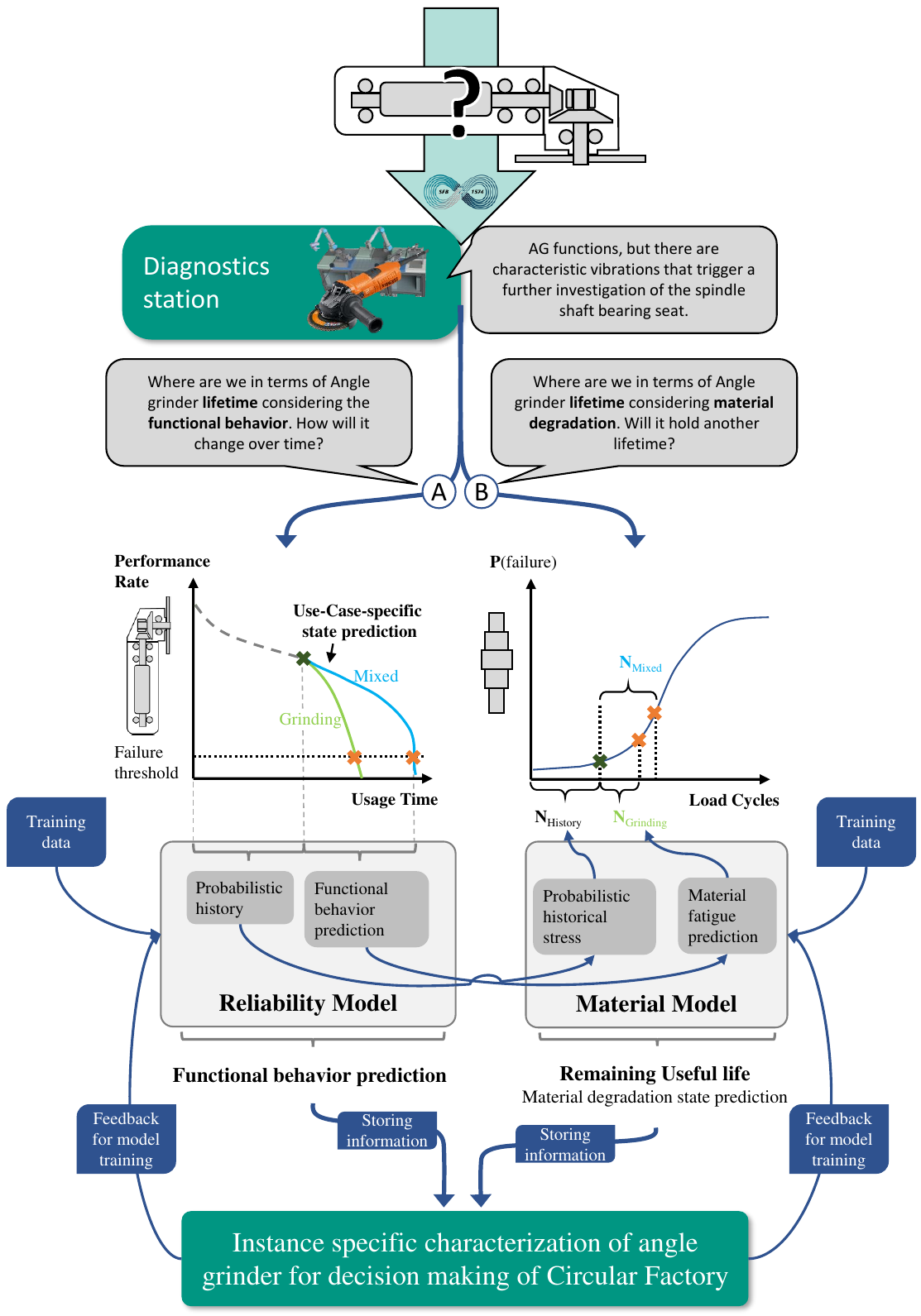}
    \caption{Angle-grinder case study used to link system-level functional behavior prediction with component-level material reliability assessment. The gear-stage response is evaluated under controlled force--torque loading, while the output shaft is considered as the fatigue-critical component for material-side reuse assessment.}
    \label{fig:caseStudy}
\end{figure}
%%%%%%%%%%%%%%%%%%%%%%%%%%%%%%%%%%%%%%%%%%%%%%%%%%%%%%%%%%%%%%%%%%%%%
\subsection{Experimental Setup for System-Functional Behavior}
\begin{figure}
    \centering
    \begin{subfigure}[t]{0.49\linewidth}
      \centering
      \includegraphics[width=\linewidth]{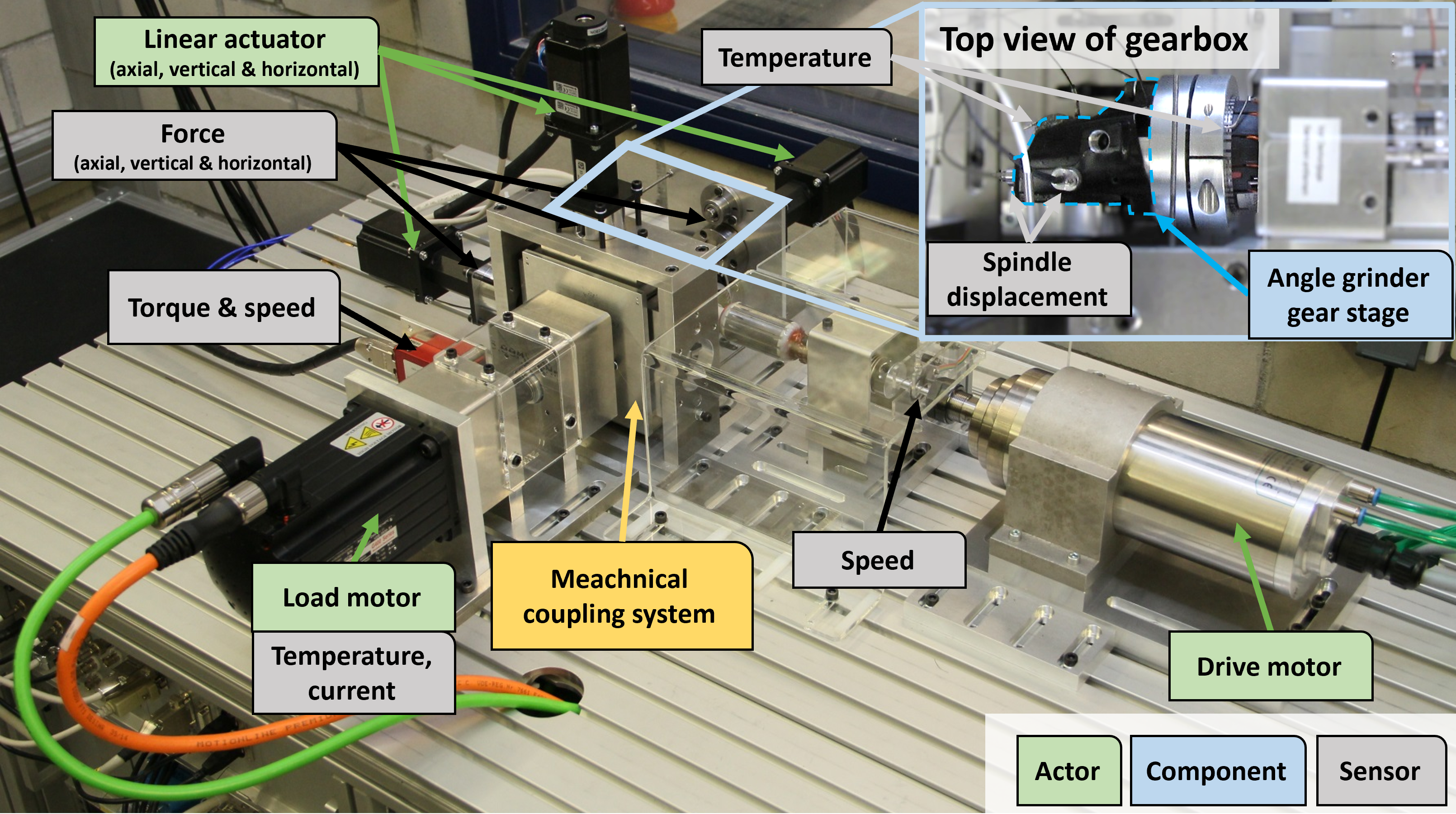}
      \caption{GKP test bench with external drive motor \cite{IGF18196N}}
      \label{fig:GKP}
    \end{subfigure}
    \hfill
    \begin{subfigure}[t]{0.49\linewidth}
      \centering
      \includegraphics[width=\linewidth]{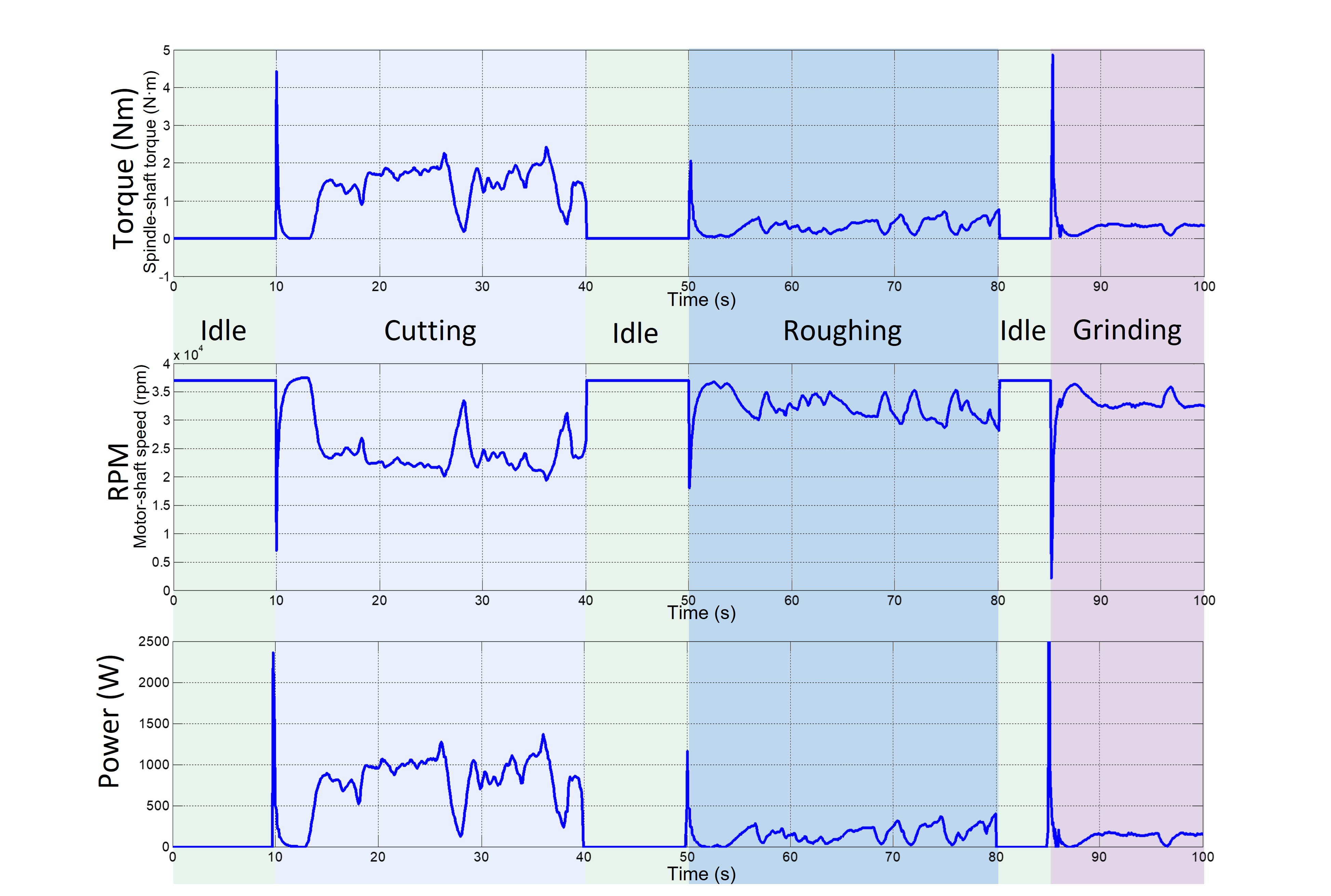}
            \caption{Repeating 100~s load cycle applied to the gear stage}
      \label{fig:test_cycle}
    \end{subfigure}
    \vspace{-4mm}
    \caption{Test bench and load cycle for functional degradation data acquisition.}
\end{figure}
%%%%%%%%%%%%%%%%%%%%%%%%%%%%%%%%%%%%%%%%%%%%%%%%%%%%%%%%%%%%%%%%%%%%%%%%%%%%%%%%%%%%
The degradation data used in this study were obtained from a previously conducted experimental study reported in \cite{IGF18196N}. The following summarizes the test bench setup and data acquisition procedure as described therein, to the extent relevant for the present modeling task.

The test bench, shown in Fig.~\ref{fig:GKP}, was designed to examine wear progression of the gear stage over 400 operating hours. Intermediate inspections were carried out after 50, 100, 200, and 300 operating hours in order to document the degradation state at defined stages of the test. The investigated subsystem was mounted on a dedicated gear-component test bench with an external drive motor and a separate load machine, with the bevel-gear stage positioned between both units. This arrangement allowed the degradation of the gear stage to be examined independently of changes in the original electric motor of the angle grinder.

A repeating load cycle of 100~s, shown in Fig.~\ref{fig:test_cycle}, was applied to reproduce representative drivetrain loading under typical operating conditions. The cycle included load segments associated with roughing, grinding, and cutting. Torque was introduced through the load machine. In addition, the spindle was loaded by three linear actuators (Nanotec L59-A), consisting of one axial actuator and two radial actuators, coupled to the output shaft through springs and a central decoupling bearing such that the linear motion was converted into one axial and two radial spindle forces.

The recorded raw data included load-side rotational speed (rpm), shaft torque (Nm) filtered with a fifth-order Butterworth low-pass filter at 30~Hz, spindle forces in the front, left, and axial directions (N), spindle displacements in the front, left, and axial directions ($\mu$m) together with lateral pinion displacement ($\mu$m), gearbox temperature ($^\circ$C), and input current and voltage (A/V). The displacement-related measurements were particularly relevant for assessing increasing mechanical play and, thus, degradation of the gear-stage system.

%%%%%%%%%%%%%%%%%%%%%%%%%%%%%%%%%%%%%%%%%%%%%%%%%%%%%%%%%%%%%%%%%%%%%%%%%%%%%%%%%%%%%%%%
\subsubsection{Variable Definition for the Angle Grinder Case Study}
For the angle grinder case study, the \textbf{state vector} contains thermal, electrical, rotational, positional, and geometric quantities, namely gearbox temperature, BLDC stator temperature, drive motor current, load speed, axial spindle position, left spindle position, front spindle position, left pinion clearance, right pinion clearance, axial spindle clearance, left spindle clearance, and front spindle clearance. These variables provide a snapshot of the thermo-mechanical condition of the tool at the forecast origin.

The \textbf{usage history} is constructed from continuously measured loading variables of the gear stage, namely the axial, left, and front spindle forces, as well as the measuring-shaft torque. These signals are physically meaningful because they directly reflect the external excitation that drives the system's thermal, electrical, and mechanical responses. They also preserve characteristic temporal patterns associated with roughing, grinding, and cutting.

The \textbf{health-indicator variables} are the pinion and spindle clearances, selected because increasing mechanical play is directly related to wear progression in the bearing and gear-support interfaces and therefore provides an interpretable representation of degradation. It should be noted that the clearance variables appear in both the state vector, as the current measured condition at the forecast anchor, and among the prediction targets, as the expected condition at the end of the subsequent usage window. The increment-based prediction formulation (Section~4) means the model learns to predict the \emph{change} in clearance over the usage window rather than reproducing the anchor value, preserving a meaningful prediction task despite this overlap.

\subsubsection{Dataset and Preprocessing}
The experimental data are stored in MATLAB files, with each file corresponding to a single operating run of the angle grinder. To avoid information leakage, the split into training, validation, and test sets is performed at the file level rather than at the window level. In the current implementation, \(80\%\) of the files are used for training, \(10\%\) for validation, and \(10\%\) for testing.

The present setup uses 12 state channels, 4 usage channels, and 9 target channels. Usage signals are processed window-wise: each force and torque channel is first denoised independently using a Savitzky--Golay filter, which smooths high-frequency noise while preserving peaks and transient changes relevant for load-history representation. Each window is then linearly interpolated to a fixed length for network input. The current configuration uses a window duration of 2.0~s, an anchor stride of 0.25~s, a history length of 4 windows, and 10 resampled points per second, yielding 20 points per usage window. Before training, all inputs and targets are standardized channel-wise using the mean and standard deviation estimated from the training set.

%%%%%%%%%%%%%%%%%%%%%%%%%%%%%%%%%%%%%%%%%%%%%%%%%%%%%%%%%%%%%%%%%%%%%%%%%%%%%%%%%%%%%%%%
\subsubsection{Model and Training Settings}
The sequence model uses a force encoder with an intermediate dimension of 64 and a force hidden dimension of 128. The state encoder uses a hidden dimension of 64, and the fused representation is mapped to a hidden dimension of 128 before entering the recurrent block. The recurrent backbone is a single-layer unidirectional LSTM with hidden dimension 128, trained in one-to-one mode so that predictions are produced at every sequence position. A dropout of 0.1 is applied in the encoder and regression head. Only the first state vector of each sequence is passed to the state encoder, while the usage history is represented by the full sequence of force- and torque-window encodings. Uncertainty estimation is enabled for all nine output variables. The resulting network contains 238{,}962 trainable parameters.

Training uses the incremental target formulation described in Section~4, with the target mode set to the median summary over the corresponding target window. The output weights are \([1.0, 1.0, 6.0, 6.0, 1.0, 1.0, 2.0, 1.0, 1.0]\), placing stronger emphasis on drive motor current, load speed, and axial spindle clearance. Optimization uses AdamW with a learning rate of \(5 \times 10^{-3}\) and weight decay \(10^{-4}\), with a cosine annealing scheduler (\(T_{\max} = 128\), minimum learning rate \(10^{-6}\)). Training uses a batch size of 16, gradient clipping of 4.0, and gradient accumulation of 8, for up to 128 epochs with a minimum of 15 epochs.

%%%%%%%%%%%%%%%%%%%%%%%%%%%%%%%%%%%%%%%%%%%%%%%%%%%%%%%%%%%%%%%%%%%%%%%%%%%%%%%%%%%%%%%%
\subsubsection{Evaluation Metrics}
Performance on the held-out test set is quantified by complementary point-prediction and reliability metrics. For a ground-truth value \(y_i\) and a prediction \(\hat{y}_i\), the tolerance-based accuracy at tolerance \(\tau\) is defined as
\begin{equation}
\mathrm{Acc}_{\tau} = \frac{1}{N}\sum_{i=1}^{N}\mathbf{1}\!\left(\left|\hat{y}_i-y_i\right| \le \tau \left|y_i\right|\right),
\end{equation}
where \(\mathbf{1}(\cdot)\) denotes the indicator function. The main reported value uses \(\tau=0.02\), corresponding to a \(2\%\) relative error tolerance. Tolerance-sensitivity curves are also obtained by varying \(\tau\) over a prescribed interval, and the area-under-curve summary is used to compare robustness across outputs.

Absolute regression quality is summarized through the mean absolute error (MAE),
%\begin{equation}
%\mathrm{MAE} = \frac{1}{N}\sum_{i=1}^{N}\left|\hat{y}_i-y_i\right|.
%\end{equation}
To make errors comparable across outputs at different physical scales, normalized metrics, such as Normalized Mean Absolute Error (NMAE) and Normalized Root Mean Squared Error (NRMSE), are reported using standardized targets and predictions.
%To make errors comparable across outputs with different physical scales, normalized metrics are reported on standardized targets \(z_i\) and predictions \(\hat{z}_i\):
%\begin{equation}
%\mathrm{NMAE} = \frac{1}{N}\sum_{i=1}^{N}\left|\hat{z}_i-z_i\right|, \qquad
%\mathrm{NRMSE} = \sqrt{\frac{1}{N}\sum_{i=1}^{N}\left(\hat{z}_i-z_i\right)^2}.
%\end{equation}
Additionally, the coefficient of determination \(R^2\) is used to quantify how well the model explains the variance of each target variable.

Reliability is assessed from the predictive Gaussian outputs. For each output variable \(m\), an exceedance threshold \(\tau_m\) is defined as the empirical mean plus \(2.5\) standard deviations of that variable over the dataset, consistent with the threshold notation used in Algorithm~\ref{alg:combined_reliability}. These thresholds require application-specific justification before use as engineering acceptance limits for redeployment. The predicted exceedance probability is obtained from the predictive mean \(\mu_i\) and standard deviation \(\sigma_i\) as
\begin{equation}
p_i = 1 - \Phi\!\left(\frac{\tau_m-\mu_i}{\sigma_i}\right),
\end{equation}
where \(\Phi(\cdot)\) denotes the standard normal cumulative distribution function. These probabilities are evaluated through four complementary views: sequence-local exceedance, Monte Carlo first-crossing, endpoint calibration, and Weibull survival analysis. Calibration quality is summarized by the Brier score,
\begin{equation}
\mathrm{Brier} = \frac{1}{N}\sum_{i=1}^{N}\left(p_i-e_i\right)^2,
\end{equation}
where \(e_i \in \{0,1\}\) denotes the observed exceedance event, together with the expected calibration error (ECE) computed from binned predicted probabilities. For the Monte Carlo first-crossing analysis, 1000 trajectories are sampled per sequence window to estimate the probability and timing of a first threshold crossing.

%%%%%%%%%%%%%%%%%%%%%%%%%%%%%%%%%%%%%%%%%%%%%%%%%%%%%%%%%%%%%%%%%%%%%%%%%%%%%%%%%%%%%%%%
\subsection{Experimental Setup for Component-Material Behavior}
The specimens were obtained from original spare parts of the angle grinder. The samples consist of output shafts, with their geometry defined by the technical drawing shown in Fig.~\ref{fig:main}.The chemical composition was measured by spark optical emission spectrometry (S-OES) and is listed in Table~\ref{tab:chemical_composition}. The examined material most closely corresponds to an alloyed quenched-and-tempered steel, AISI~4130 (25CrMo4). Microstructural characterization reveals that the material is in a tempered martensitic state, as the investigation was carried out within the framework of a reverse engineering study.
%(generation~\textbf{[XX]}, manufactured by C.\ \& E.\ Fein GmbH)
\subsubsection{Mechanical Fatigue Testing}
\begin{figure}
    \centering
    \begin{subfigure}[t]{0.39\linewidth}
        \centering
        \includegraphics[width=\linewidth, height=0.27\textheight, keepaspectratio,
            trim=0.4cm 0.4cm 0.4cm 0.4cm, clip]{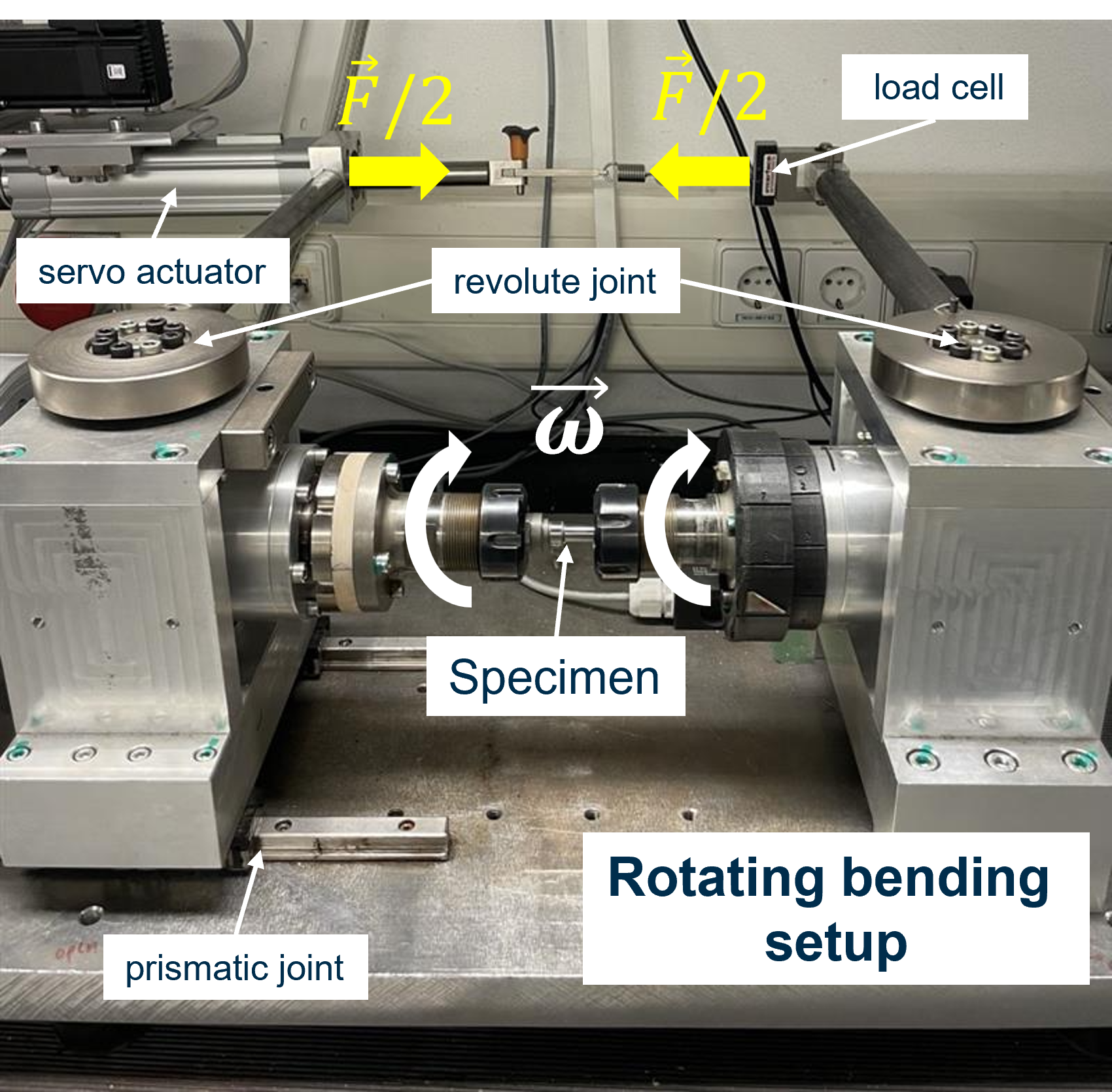}
        \caption{Rotating bending test setup with specimen mounted in the testing machine}
        \label{fig:rotating-bending-setup}
    \end{subfigure}
    \hfill
    \begin{subfigure}[t]{0.54\linewidth}
        \centering
        \includegraphics[width=\linewidth,
            trim=0cm 0.cm 0.cm 0.cm, clip]{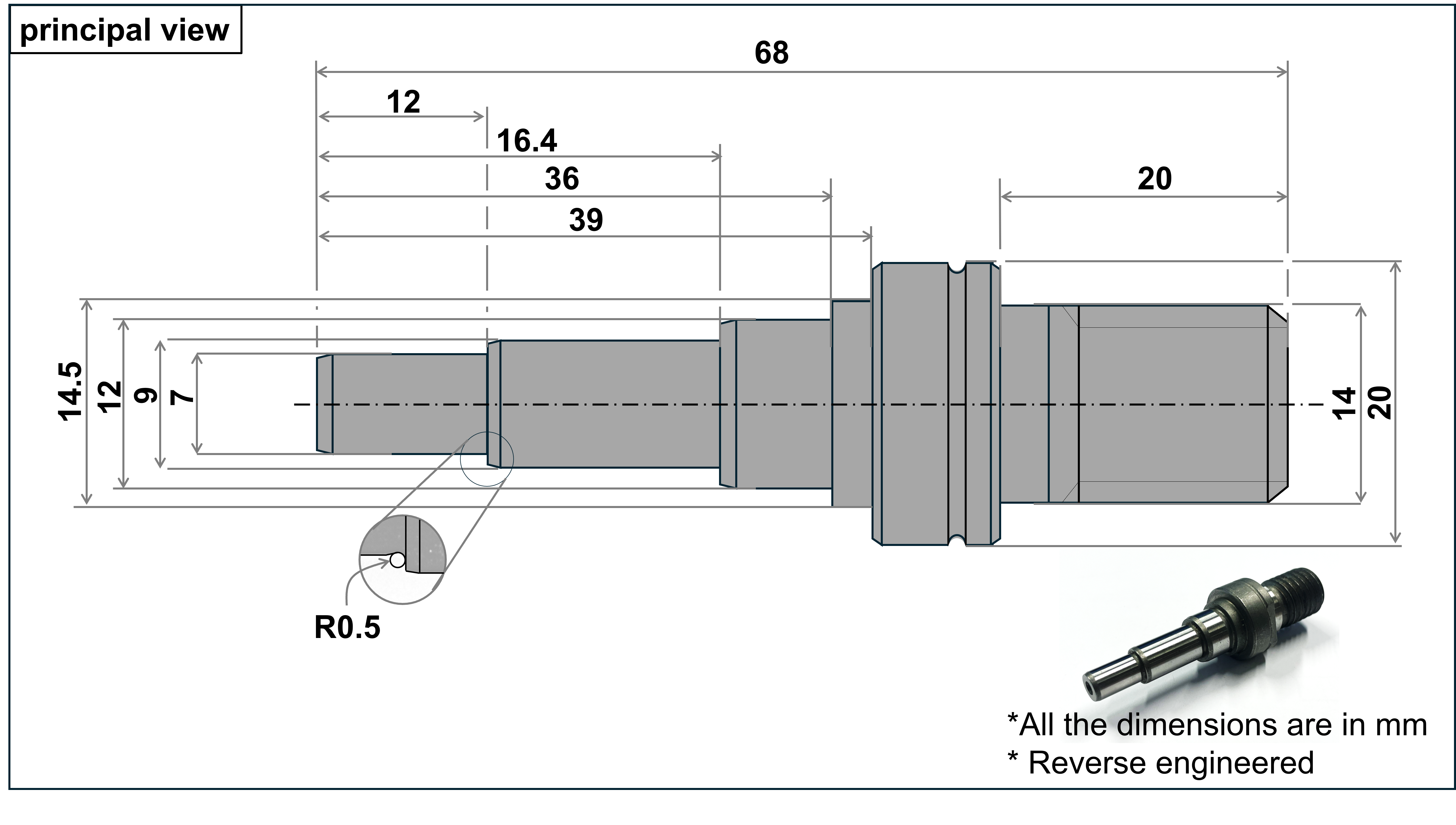}
        \caption{Technical drawing of the output shaft specimen}
        \label{fig:shaft-drawing}
    \end{subfigure}
    \caption{Rotating bending test setup and output shaft specimen geometry.}
    \label{fig:main}
\end{figure}

Fatigue behavior was investigated using a moment-controlled rotating bending test system, shown in Fig.~\ref{fig:main}. All tests were performed at a constant frequency of 50~Hz under fully reversed loading conditions ($R = -1$). The S--N curve was determined by combining a staircase method in accordance with DIN~50100, using at least 20 specimens, with a horizon approach employing at least four specimens per stress amplitude level to ensure adequate statistical representation across the investigated load range.

\begin{table}[pos=h]
\caption{Chemical composition of the output shaft material (AISI~4130 / 25CrMo4) in wt-\%, determined by optical emission spectroscopy (OES).}
\label{tab:chemical_composition}
\centering
\begin{tabular}{llllllllll}
\toprule
Element & C & Si & Mn & P & S & Cu & Cr & Ni & Mo \\
\midrule
        & 0.272 & 0.270 & 0.507 & 0.015 & 0.004 & 0.025 & 0.866 & 0.019 & 0.184 \\
\bottomrule
\end{tabular}
\end{table}

\subsubsection{Finite Element Stress Characterization}
\begin{figure}
    \centering
    \begin{subfigure}[t]{0.49\linewidth}
        \centering
        \includegraphics[width=\linewidth,
            trim=1.7cm 0.4cm 0.4cm 0.4cm, clip]{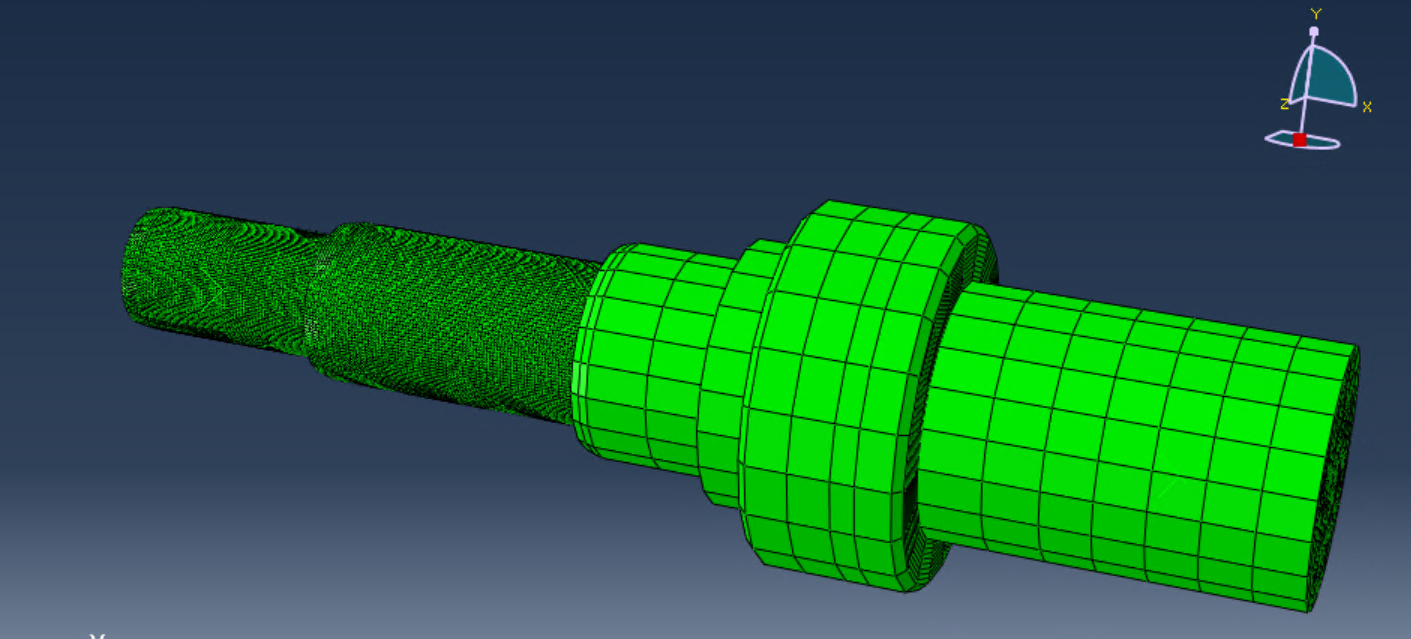}
        \caption{Finite element mesh of the output shaft with local refinement at the notch}
        \label{fig:fem-mesh}
    \end{subfigure}
    \hfill
    \begin{subfigure}[t]{0.49\linewidth}
        \centering
        \includegraphics[width=0.9\linewidth,
            trim=0.4cm 0.4cm 0.4cm 0.4cm, clip]{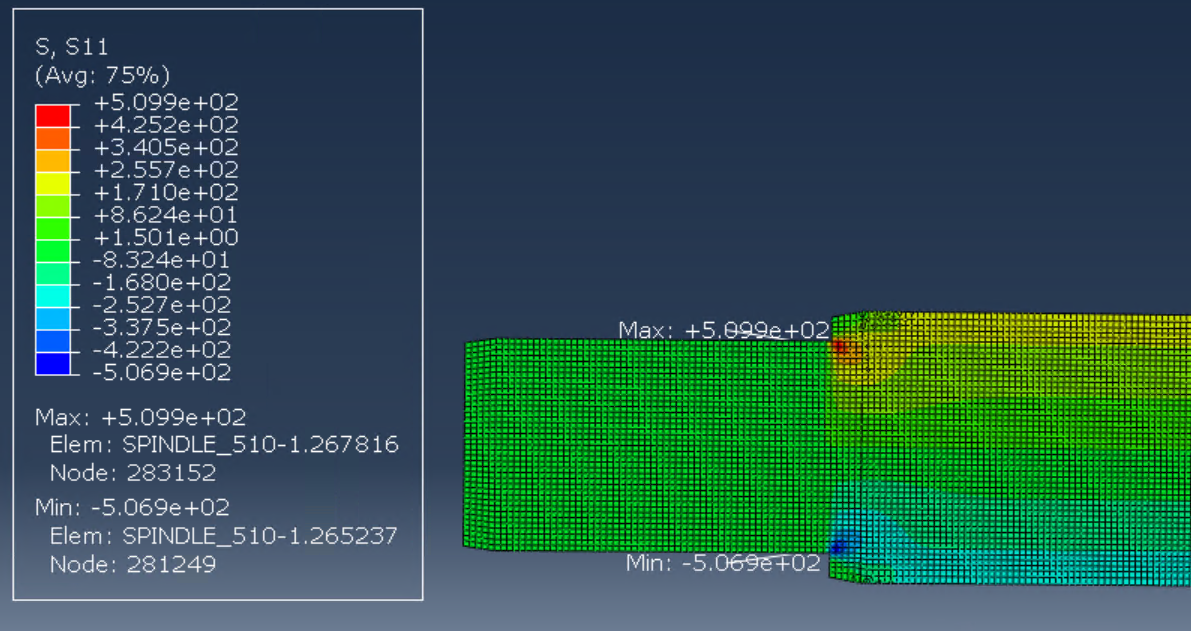}
        \caption{Stress distribution under rotating bending, showing peak stress concentration at the critical notch}
        \label{fig:fem-results}
    \end{subfigure}
    \vspace{-2mm}
    \caption{Finite element model and stress field of the output shaft under rotating bending. The notch root stress serves as the governing fatigue parameter in the material assessment.}
    \label{fig:FEM-shaft}
\end{figure}

Finite element analysis was carried out in Abaqus to determine the local stress distribution in the output shaft under rotating bending and to support the interpolation-based stress reconstruction described in Section~4. For the rotating bending configuration, a three-dimensional model was constructed based on the shaft technical drawing. The mesh was locally refined in the notch region to resolve the stress gradient accurately, and the analysis was performed under linear-elastic material assumptions. The resulting stress field, shown in Fig.~\ref{fig:FEM-shaft}, confirms a pronounced stress concentration at the notch root, with the peak local notch stress used as the governing parameter for all fatigue and crack-propagation calculations.

\subsubsection{Hardness and Microstructure}
To characterize the hardness distribution and microstructural state of the shaft, specimens were sectioned in the radial direction ($XZ$ plane). Hardness mapping was performed using a Qness Q10 microhardness tester with a test load of 1~N and a dwell time of 10~s. Measurements were conducted in the notch region at the smallest shaft diameter, with indentation lines extending from the surface toward the center. Mean values were calculated from three independent depth profiles to obtain a representative hardness profile and reduce local scatter. The specimens for microstructural characterization were prepared by cutting, embedding, grinding to 2400 grit, polishing.
%%%%%%%%%%%%%%%%%%%%%%%%%%%%%%%%%%%%%%%%%%%%%%%%%%%%%%%%%%%%%%%%%%%%%%%%%%%%
%%%%%%%%%%%%%%%%%%%%%%%%%%%%%%%%%%%%%%%%%%%%%%%%%%%%%%%%%%%%%%%%%%%%%%%%%%%
\section{Results}\label{}
\subsection{System--Functional Behavior Results}
\subsubsection{Ablation Study on Usage-History Inputs}
To better understand how the choice of usage-history inputs influences the learned predictor, four auxiliary ablation runs were compared. Across these runs, the training and evaluation protocols were kept fixed, while only the channels used to describe the recent usage history were varied. The purpose of this comparison is to assess whether the model requires sufficiently representative load information to learn the correct functional response.

Table~\ref{tab:input_ablation} reports the held-out \(2\%\)-tolerance accuracy for each run. This metric is used as a model-comparison criterion rather than as a failure threshold; it measures the fraction of predictions whose relative error is within \(\pm 2\%\) of the reference value. The delta columns are computed relative to the axial-force-only baseline. Adding more representative usage-history information improves mean accuracy from 0.8965 (axial force only) to 0.9652 (all four channels), a gain of \(+0.0687\). The strongest sensitivity is observed in the two outputs most directly linked to drivetrain loading: drive motor current and load speed (CAN). Adding the measuring-shaft torque alone yields a substantially larger gain for these outputs (\(+0.2172\) and \(+0.3165\)) than adding the two additional spindle-force channels alone (\(+0.1158\) and \(+0.1194\)). The best configuration, torque combined with the full spindle-force set, reaches accuracies of 0.9174 and 0.8675 for these two variables, with the largest deltas of \(+0.2438\) and \(+0.3397\). By contrast, the geometry-related axial spindle clearance is less sensitive to torque: it already reaches 0.8984 with axial force only and shows a small negative delta of \(-0.0038\) when torque alone is added, indicating that force signals already capture the mechanically induced response for this variable. The two thermal variables remained effectively unchanged across all four runs.
\begin{figure}
    \centering
    \begin{subfigure}[t]{0.45\linewidth}
        %\centering
        \includegraphics[width=\linewidth,
            trim=0.4cm 0.4cm 0.4cm 0.4cm, clip]{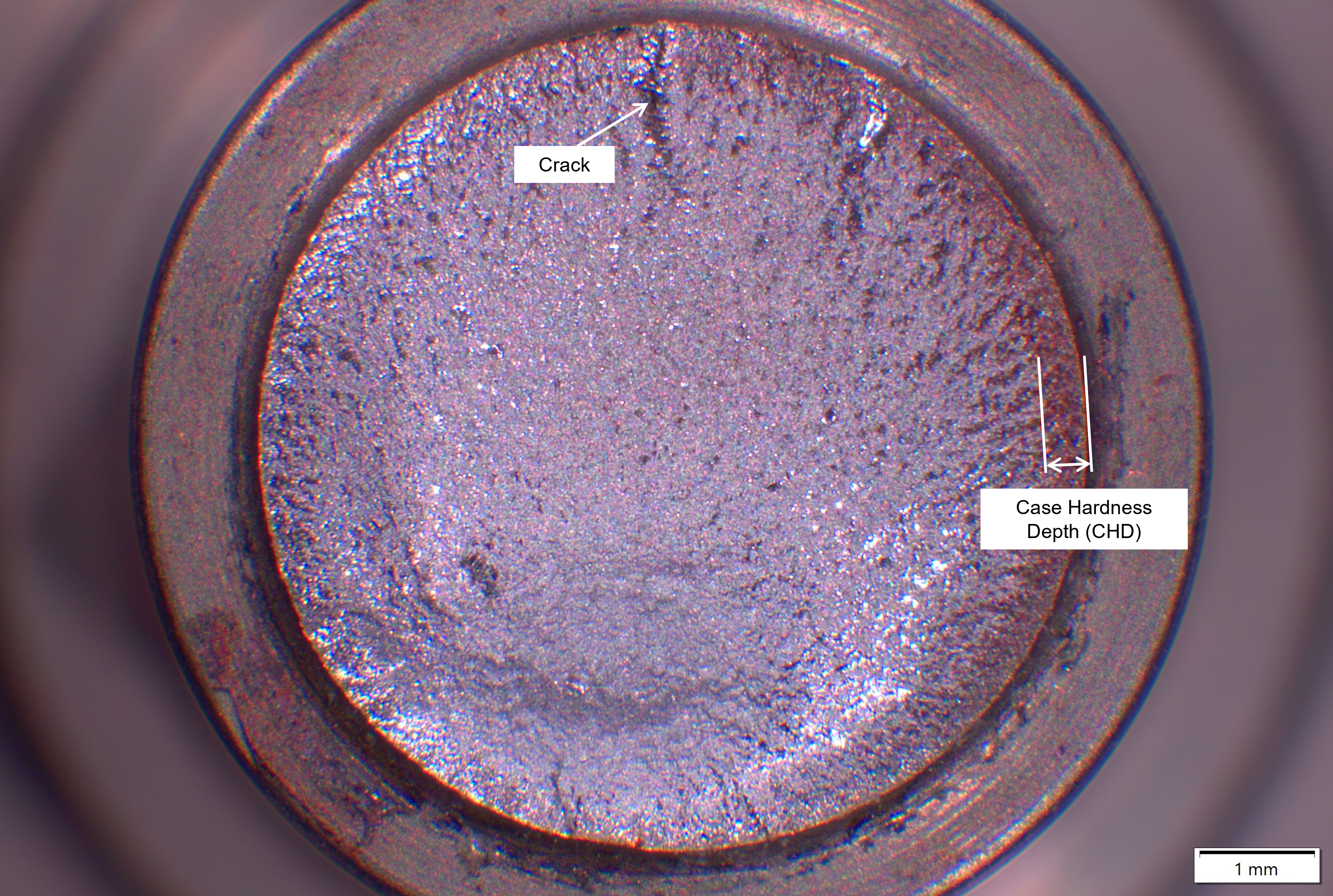}
        \caption{Fractography of the shaft fracture surface showing a surface-initiated fatigue crack and the case-hardened layer (CHD); the crack depth relative to the hardened zone defines the critical crack size $a_c$ and constrains crack propagation, thereby governing the remaining useful life (RUL)}
        \label{fig:hardness-map}
    \end{subfigure}
    %\hfill
    \begin{subfigure}[t]{0.5\linewidth}
        %\centering
        \includegraphics[width=0.9\linewidth,
            trim=0.4cm 0.4cm 0.4cm 0.4cm, clip]{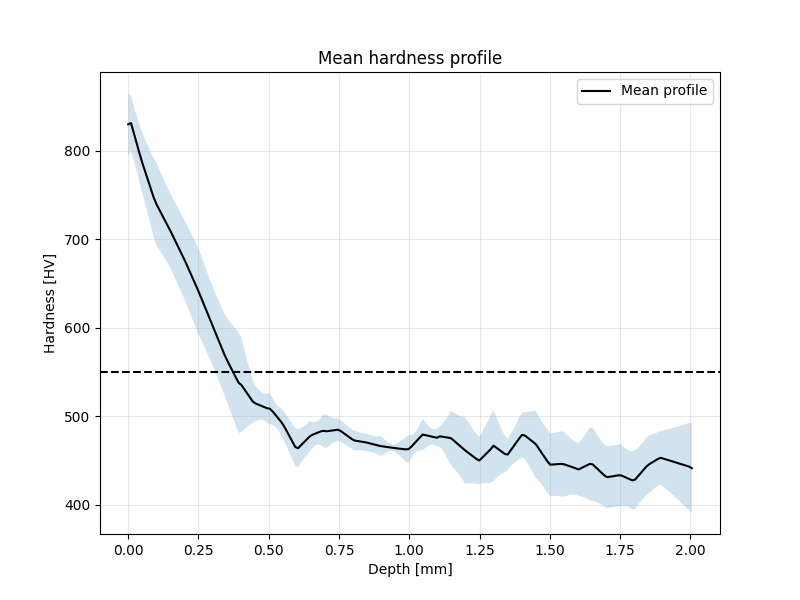}
        \caption{In-depth hardness profile from surface to core}
        \label{fig:hardness-profile}
    \end{subfigure}
    \caption{Hardness characterization of the output shaft at the critical notch. The effective case-hardened depth of approximately 0.369~mm, determined in accordance with ISO~2639, defines the upper limit for crack propagation used.}
    \label{fig:hardness}
\end{figure}

\begin{table}[t]
\centering
\scriptsize
\caption{Ablation of the usage-history inputs. A checkmark indicates that the corresponding channel is included in the recent usage-history representation. Reported results are held-out \(2\%\)-tolerance accuracies. Each delta value is computed relative to the axial-force-only baseline.}
\label{tab:input_ablation}
\resizebox{\linewidth}{!}{%
\begin{tabular}{cccccccccccc}
\hline
\makecell{Axial\\Spindle Force} & \makecell{Left\\Spindle Force} & \makecell{Front\\Spindle Force} & \makecell{Measuring-Shaft\\Torque} & Mean Acc. & \makecell{\(\Delta\) Mean\\Acc.} & \makecell{Drive Motor\\Current} & \makecell{\(\Delta\) Drive Motor\\Current} & \makecell{Load Speed\\(CAN)} & \makecell{\(\Delta\) Load Speed\\(CAN)} & \makecell{Axial Spindle\\Clearance} & \makecell{\(\Delta\) Axial Spindle\\Clearance} \\
\hline
\(\checkmark\) &  &  &  & 0.8965 & +0.0000 & 0.6736 & +0.0000 & 0.5278 & +0.0000 & 0.8984 & +0.0000 \\
\(\checkmark\) & \(\checkmark\) & \(\checkmark\) &  & 0.9267 & +0.0302 & 0.7894 & +0.1158 & 0.6472 & +0.1194 & \textbf{0.9264} & +0.0280 \\
\(\checkmark\) &  &  & \(\checkmark\) & 0.9572 & +0.0607 & 0.8908 & +0.2172 & 0.8443 & +0.3165 & 0.8946 & -0.0038 \\
\(\checkmark\) & \(\checkmark\) & \(\checkmark\) & \(\checkmark\) & \textbf{0.9652} & \textbf{+0.0687} & \textbf{0.9174} & \textbf{+0.2438} & \textbf{0.8675} & \textbf{+0.3397} & 0.9124 & +0.0140 \\
\hline
\end{tabular}%
}
\end{table}
%%%%%%%%%%%%%%%%%%%%%%%%%%%%%%%%%%%%%%%%%%%%%%%%%%%%%%%%%%%%%%%%%%%%%%%%%%%%%%%%%%%%%%%%
\subsubsection{Ablation Study on Recurrent Backbone}
To assess the effect of the recurrent backbone, the selected best held-out runs for LSTM, GRU, and xLSTM were compared under otherwise matched settings. Table~\ref{tab:rnn_ablation} shows that the LSTM achieved the strongest overall performance: mean accuracy 0.9652, mean NRMSE 0.0297, and mean \(R^2\) 0.8365. The corresponding GRU values were 0.9516, 0.0364, and 0.8262, while the xLSTM values were 0.9448, 0.0362, and 0.8250.

The clearest differences appeared in drive motor current and load speed (CAN). The LSTM reached \(2\%\)-tolerance accuracies of 0.9174 and 0.8675 with \(R^2\) values of 0.9750 and 0.9924. The GRU reached 0.8768 and 0.7994 with \(R^2\) values of 0.9616 and 0.9875, and the xLSTM reached 0.8697 and 0.7524 with \(R^2\) values of 0.9605 and 0.9887. The xLSTM remained competitive with GRU on a few secondary metrics but did not surpass the LSTM on any of the primary outputs.

\begin{table}[h]
\centering
\scriptsize
\caption{Ablation of the recurrent layer type. Reported values are held-out \(2\%\)-tolerance accuracies and complementary regression metrics for the three recurrent backbones under otherwise matched settings.}
\label{tab:rnn_ablation}
\resizebox{0.75\linewidth}{!}{%
\begin{tabular}{clccccc}
\hline
\multicolumn{1}{l}{Recurrent Layer type} & Output variable         & Accuracy & MAE & NMAE & NRMSE & \(R^2\) \\ \hline
\multirow{9}{*}{LSTM}                    & Gearbox Temperature     & 1.0000 & 0.0260 & 0.0027 & 0.0036 & 0.9999 \\
                                         & BLDC Stator Temperature & 1.0000 & 0.0239 & 0.0016 & 0.0022 & 0.9999 \\
                                         & Drive Motor Current     & 0.9174 & 0.0200 & 0.0890 & 0.1435 & 0.9750 \\
                                         & Load Speed (CAN)        & 0.8675 & 61.4052 & 0.0483 & 0.0834 & 0.9924 \\
                                         & Left Pinion Clearance   & 0.9983 & 0.0053 & 0.0053 & 0.0069 & 0.7557 \\
                                         & Right Pinion Clearance  & 1.0000 & 0.0032 & 0.0032 & 0.0042 & 0.3381 \\
                                         & Axial Spindle Clearance & 0.9124 & 0.0088 & 0.0088 & 0.0126 & 0.8756 \\
                                         & Left Spindle Clearance  & 0.9998 & 0.0033 & 0.0033 & 0.0044 & 0.9306 \\
                                         & Front Spindle Clearance & 0.9912 & 0.0050 & 0.0050 & 0.0068 & 0.6616 \\ \hline
\multirow{9}{*}{GRU}                     & Gearbox Temperature     & 1.0000 & 0.0264 & 0.0027 & 0.0037 & 0.9999 \\
                                         & BLDC Stator Temperature & 1.0000 & 0.0270 & 0.0019 & 0.0025 & 0.9999 \\
                                         & Drive Motor Current     & 0.8768 & 0.0247 & 0.1096 & 0.1779 & 0.9616 \\
                                         & Load Speed (CAN)        & 0.7994 & 79.0138 & 0.0621 & 0.1074 & 0.9875 \\
                                         & Left Pinion Clearance   & 0.9934 & 0.0054 & 0.0054 & 0.0071 & 0.7403 \\
                                         & Right Pinion Clearance  & 1.0000 & 0.0033 & 0.0033 & 0.0042 & 0.3198 \\
                                         & Axial Spindle Clearance & 0.9065 & 0.0091 & 0.0091 & 0.0135 & 0.8582 \\
                                         & Left Spindle Clearance  & 0.9991 & 0.0034 & 0.0034 & 0.0045 & 0.9258 \\
                                         & Front Spindle Clearance & 0.9891 & 0.0051 & 0.0051 & 0.0070 & 0.6425 \\ \hline
\multirow{9}{*}{xLSTM}                   & Gearbox Temperature     & 1.0000 & 0.0311 & 0.0032 & 0.0044 & 0.9999 \\
                                         & BLDC Stator Temperature & 1.0000 & 0.0298 & 0.0021 & 0.0028 & 0.9999 \\
                                         & Drive Motor Current     & 0.8697 & 0.0258 & 0.1146 & 0.1803 & 0.9605 \\
                                         & Load Speed (CAN)        & 0.7524 & 86.4268 & 0.0679 & 0.1019 & 0.9887 \\
                                         & Left Pinion Clearance   & 0.9907 & 0.0055 & 0.0055 & 0.0072 & 0.7333 \\
                                         & Right Pinion Clearance  & 1.0000 & 0.0033 & 0.0033 & 0.0042 & 0.3155 \\
                                         & Axial Spindle Clearance & 0.9019 & 0.0091 & 0.0091 & 0.0130 & 0.8683 \\
                                         & Left Spindle Clearance  & 0.9998 & 0.0035 & 0.0035 & 0.0046 & 0.9222 \\
                                         & Front Spindle Clearance & 0.9884 & 0.0051 & 0.0051 & 0.0071 & 0.6369 \\ \hline
\end{tabular}%
}
\end{table}
%%%%%%%%%%%%%%%%%%%%%%%%%%%%%%%%%%%%%%%%%%%%%%%%%%%%%%%%%%%%%%%%%%%%%%%%%%%%%%
\subsubsection{Held-Out Regression and Classification Performance}
Table~\ref{tab:test_metrics_full} reports the full set of point-prediction metrics for the selected LSTM run on the held-out test set. The two thermal variables, gearbox temperature and BLDC stator temperature, achieved accuracies of 1.0000 and \(R^2\) values of 0.9999. Several clearance variables were also predicted with high local precision, including left spindle clearance at 0.9998 accuracy and \(R^2=0.9306\). Drive motor current and load speed (CAN) yielded the lowest strict-tolerance accuracies, 0.9174 and 0.8675, while their \(R^2\) values remained high at 0.9750 and 0.9924.

Load speed (CAN) shows a notable pattern: it achieved the lowest \(2\%\) accuracy yet the second highest \(R^2\), indicating that its global trajectory was learned well while some local deviations exceed the narrow tolerance band. Axial spindle clearance shows the inverse: its accuracy of 0.9124 is modest, but its NRMSE of 0.0126 and \(R^2=0.8756\) indicate that absolute errors remain small. Right pinion clearance achieved an accuracy of 1.0000 alongside a low \(R^2\) of 0.3381, reflecting limited variance of this variable in the held-out set rather than large prediction error.

\begin{figure}
\centering
\begin{subfigure}[t]{0.49\textwidth}
\centering
\includegraphics[width=0.9\linewidth]{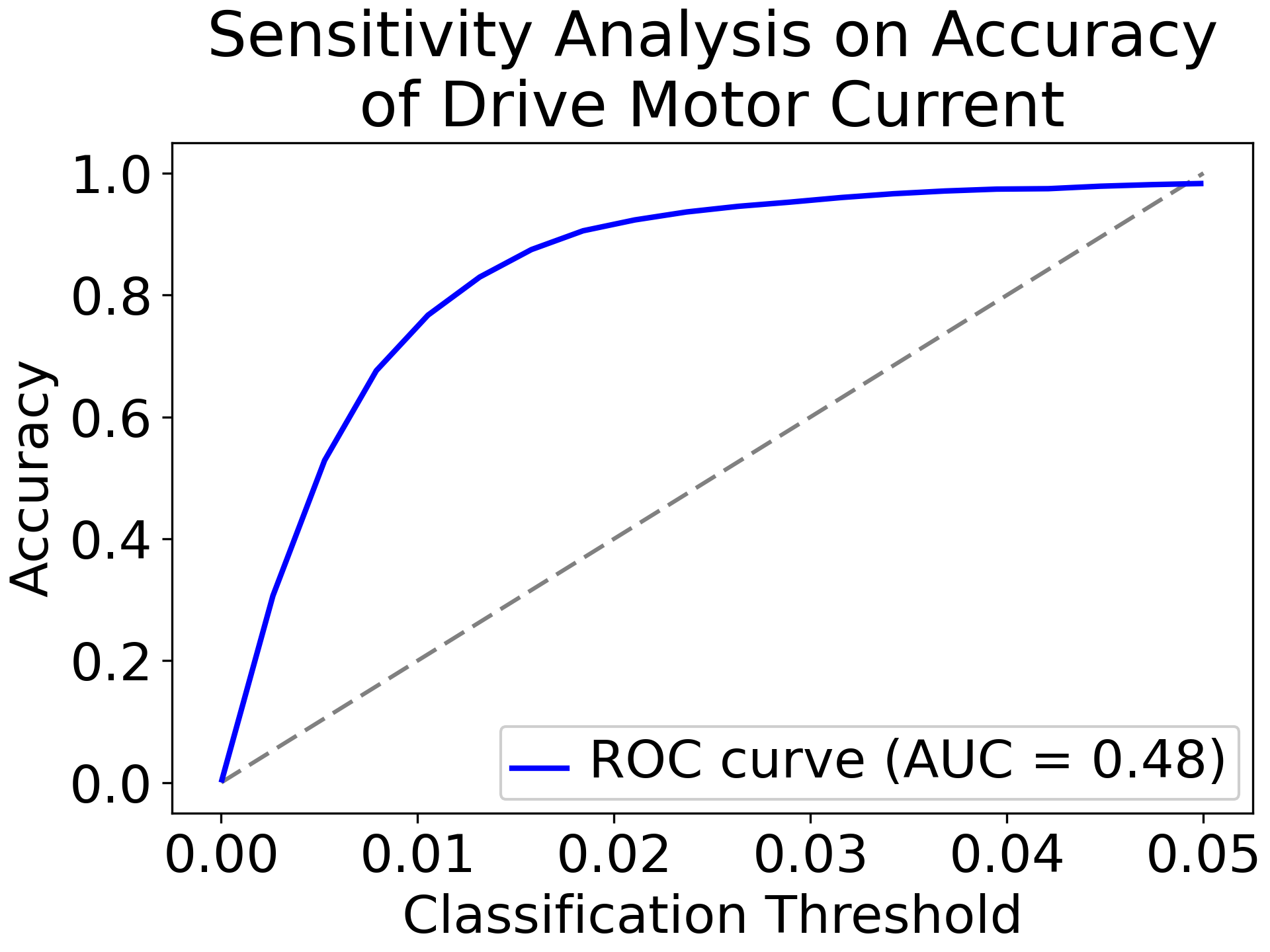}
\caption{Drive Motor Current.}
\end{subfigure}
\hfill
\begin{subfigure}[t]{0.49\textwidth}
\centering
\includegraphics[width=0.9\linewidth]{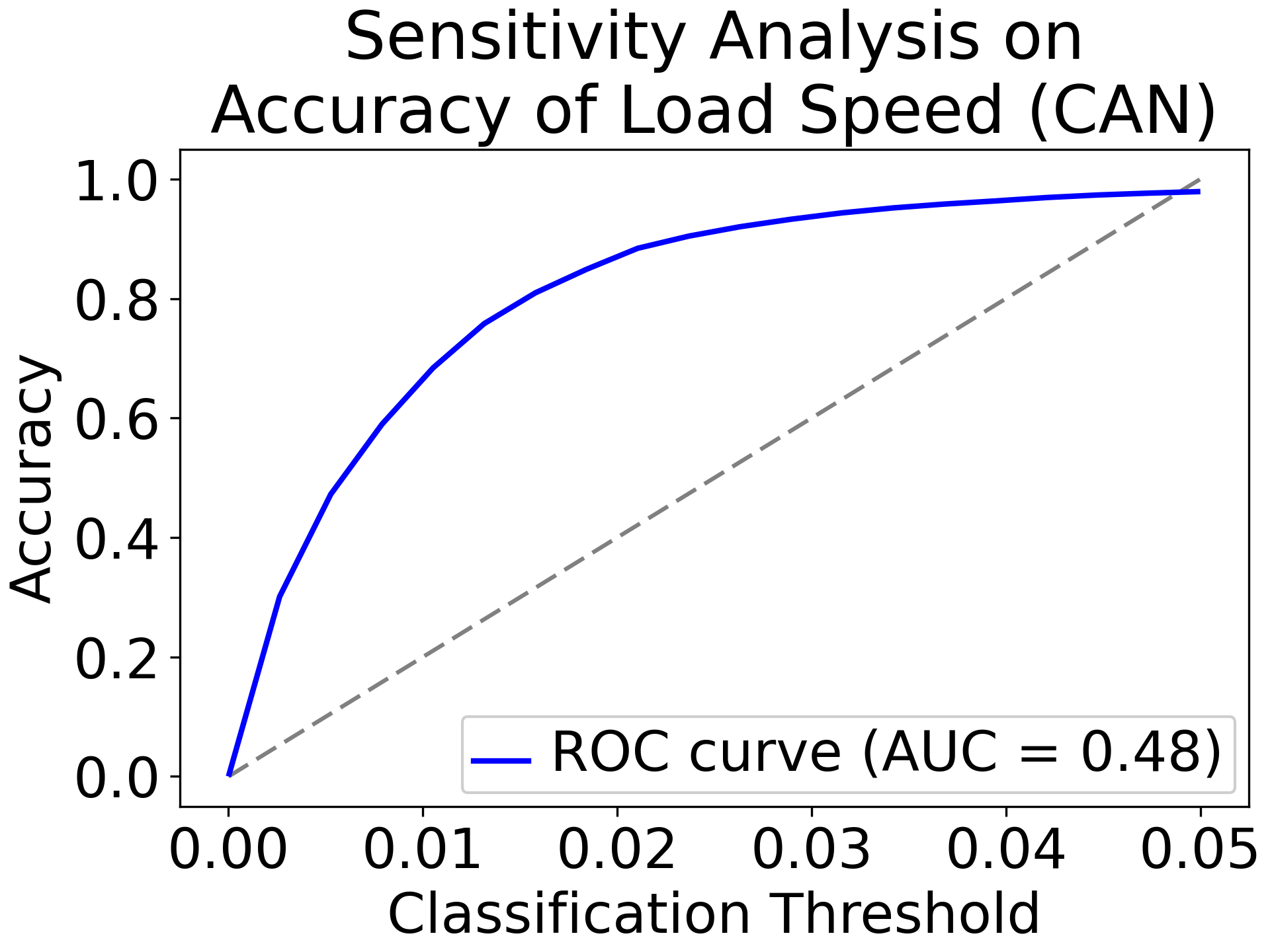}
\caption{Load Speed (CAN).}
\end{subfigure}
\caption{Tolerance-sensitivity curves for drive motor current and load speed (CAN), the two outputs with the lowest strict \(2\%\) accuracy.}
\label{fig:sensitivity_worst_two}
\end{figure}

Figure~\ref{fig:sensitivity_worst_two} shows the tolerance-sensitivity curves for drive motor current and load speed (CAN), the two outputs with the lowest strict-tolerance accuracy. Both curves rise more gradually than the remaining variables and yield a curve-area value of 0.48, confirming their greater sensitivity to tight local tolerances.

\begin{table}[t]
\centering
\scriptsize
\caption{Held-out regression metrics for the nine predicted outputs. Accuracy is defined at a \(2\%\) relative error tolerance.}
\label{tab:test_metrics_full}
\resizebox{0.75\linewidth}{!}{%
\begin{tabular}{lccccc}
\hline
Output variable & Accuracy & MAE  & NMAE & NRMSE & \(R^2\) \\
\hline
Gearbox Temperature     & 1.0000 & 0.0260 & 0.0027 & 0.0036 & 0.9999 \\
BLDC Stator Temperature & 1.0000 & 0.0239 & 0.0016 & 0.0022 & 0.9999 \\
Drive Motor Current     & 0.9174 & 0.0200 & 0.0890 & 0.1435 & 0.9750 \\
Load Speed (CAN)        & 0.8675 & 61.4052 & 0.0483 & 0.0834 & 0.9924 \\
Left Pinion Clearance   & 0.9983 & 0.0053 & 0.0053 & 0.0069 & 0.7557 \\
Right Pinion Clearance  & 1.0000 & 0.0032 & 0.0032 & 0.0042 & 0.3381 \\
Axial Spindle Clearance & 0.9124 & 0.0088 & 0.0088 & 0.0126 & 0.8756 \\
Left Spindle Clearance  & 0.9998 & 0.0033 & 0.0033 & 0.0044 & 0.9306 \\
Front Spindle Clearance & 0.9912 & 0.0050 & 0.0050 & 0.0068 & 0.6616 \\
\hline
\end{tabular}%
}
\end{table}

%%%%%%%%%%%%%%%%%%%%%%%%%%%%%%%%%%%%%%%%%%%%%%%%%%%%%%%%%%%%%%%%%%%%%%%%%%%%%%%%%%%%%%%%
\subsection{Reliability Results}
The held-out run was evaluated from a reliability perspective using four complementary views: sequence-local exceedance, Monte Carlo first-crossing, endpoint calibration, and Weibull survival analysis. Exceedance thresholds were defined as the empirical mean plus \(2.5\) standard deviations for each output variable. Among the nine outputs, drive motor current was the only variable with clearly non-negligible exceedance event support. Its mean sequence-window failure probability was 0.0600, compared with an observed window failure rate of 0.0646. The Monte Carlo first-crossing estimate was also 0.0600, and the mean endpoint risk was 0.0154 against an observed endpoint exceedance rate of 0.0161. The window ECE values were 0.0085 and 0.0086 for the sequence-local and Monte Carlo views, and the endpoint ECE was 0.0033. All remaining outputs stayed effectively in the non-failure regime.

\begin{figure}
\centering
\begin{subfigure}[t]{0.49\textwidth}
\centering
\includegraphics[width=\linewidth]{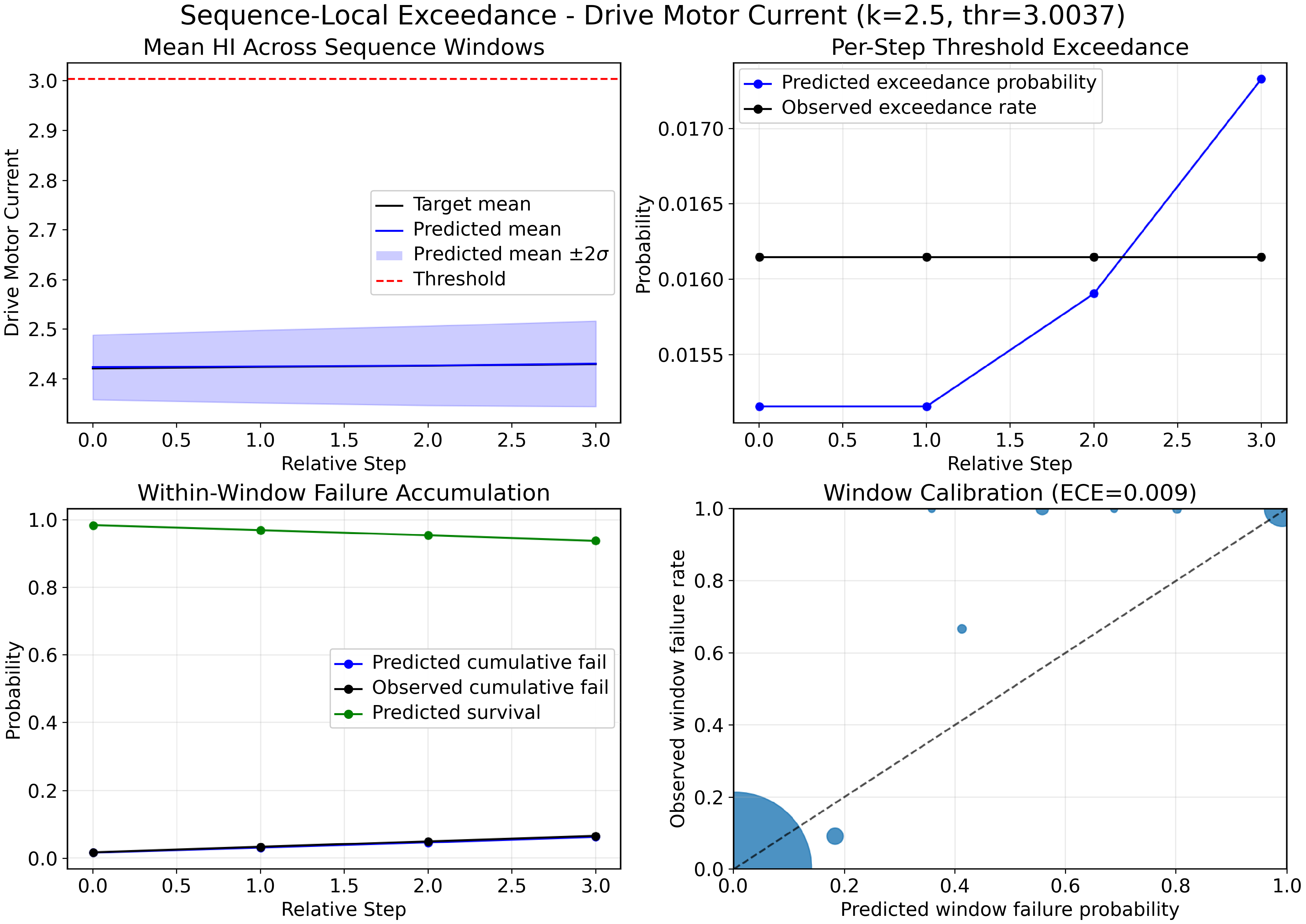}
\caption{Drive Motor Current.}
\end{subfigure}
\hfill
\begin{subfigure}[t]{0.49\textwidth}
\centering
\includegraphics[width=\linewidth]{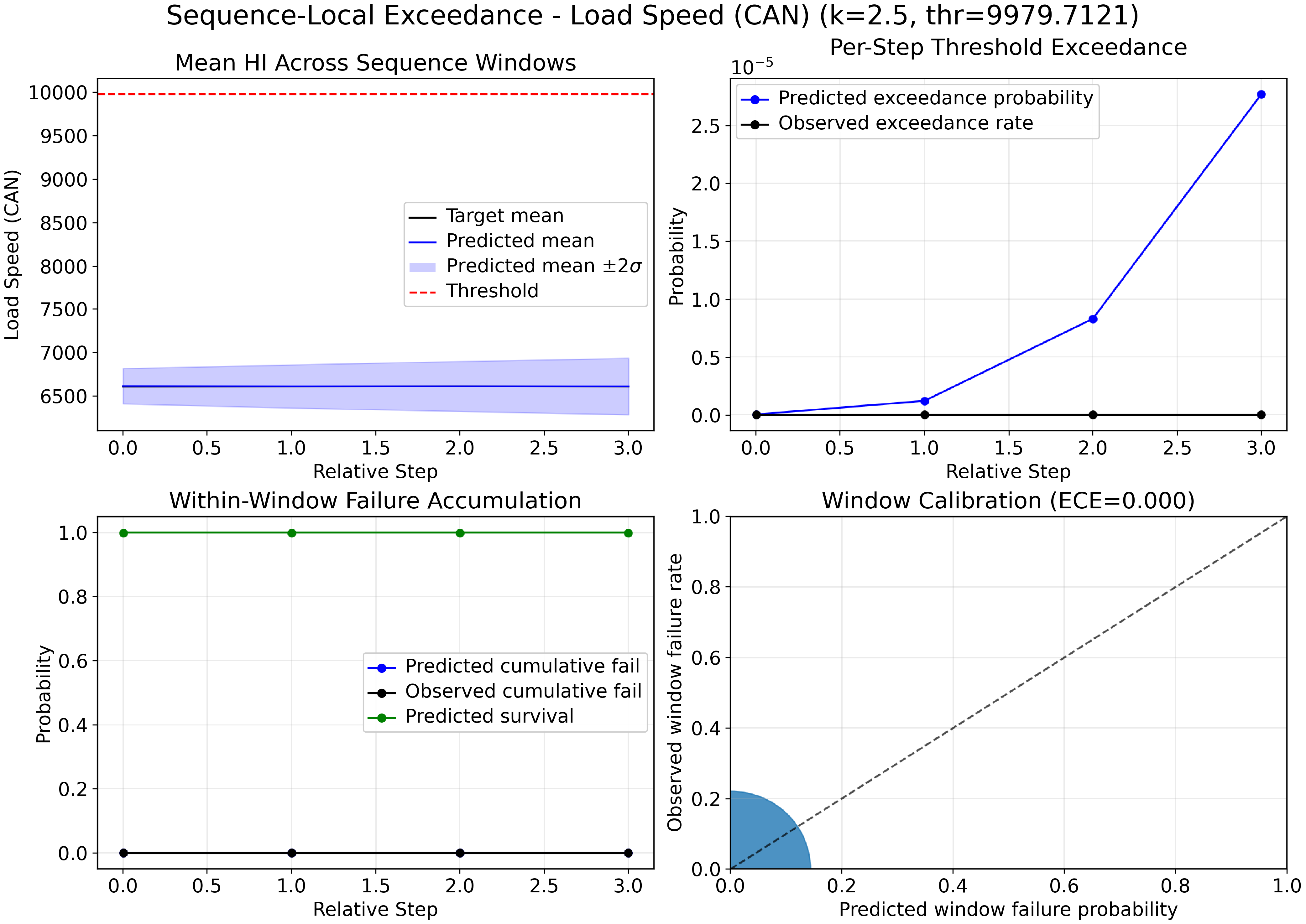}
\caption{Load Speed (CAN).}
\end{subfigure}
\caption{Sequence-local exceedance analysis for drive motor current and load speed (CAN) on the held-out run.}
\label{fig:reliability_sequence_representative}
\end{figure}

\begin{figure}
\centering
\begin{subfigure}[t]{0.49\textwidth}
\centering
\includegraphics[width=\linewidth]{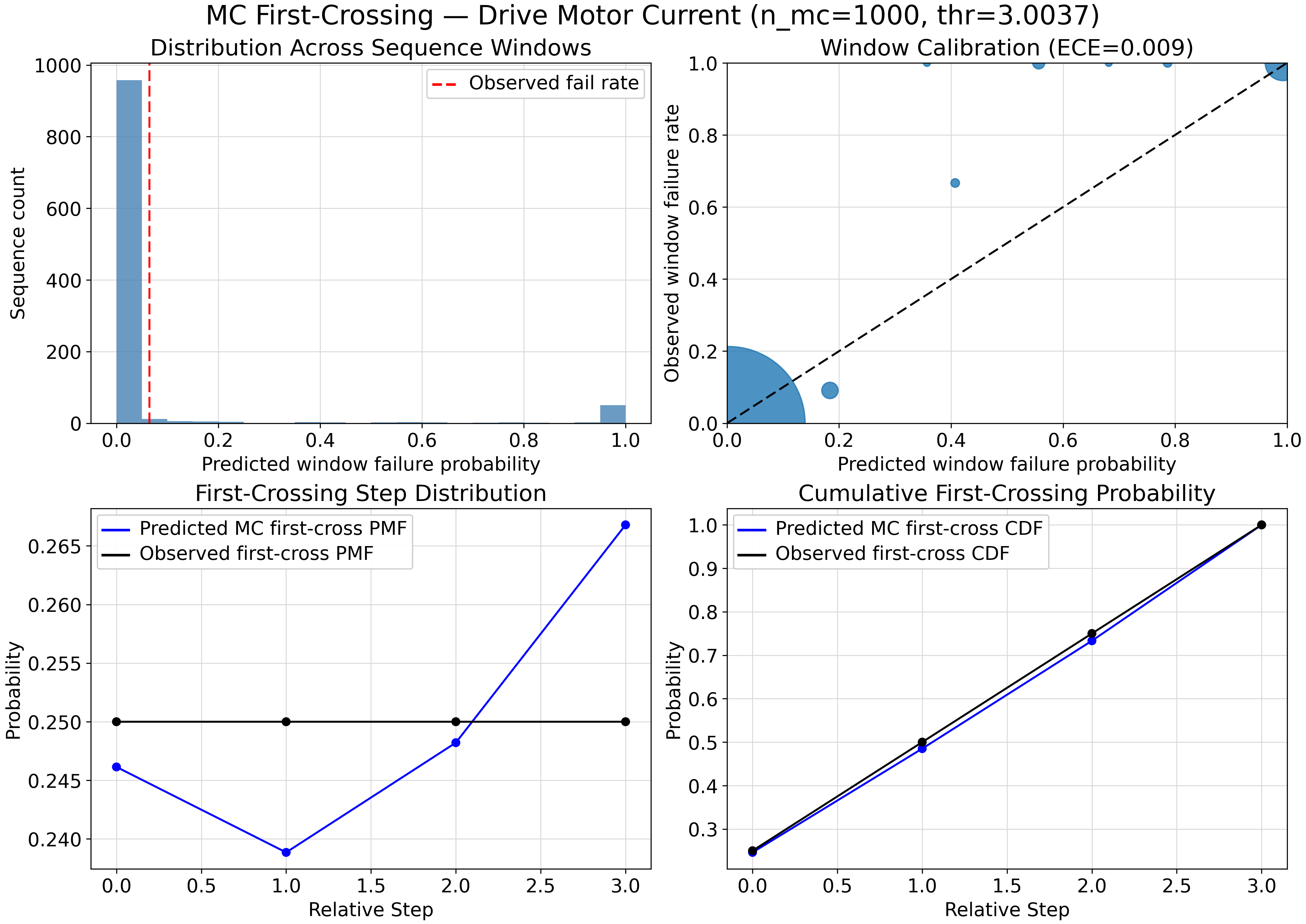}
\caption{Drive Motor Current.}
\end{subfigure}
\hfill
\begin{subfigure}[t]{0.49\textwidth}
\centering
\includegraphics[width=\linewidth]{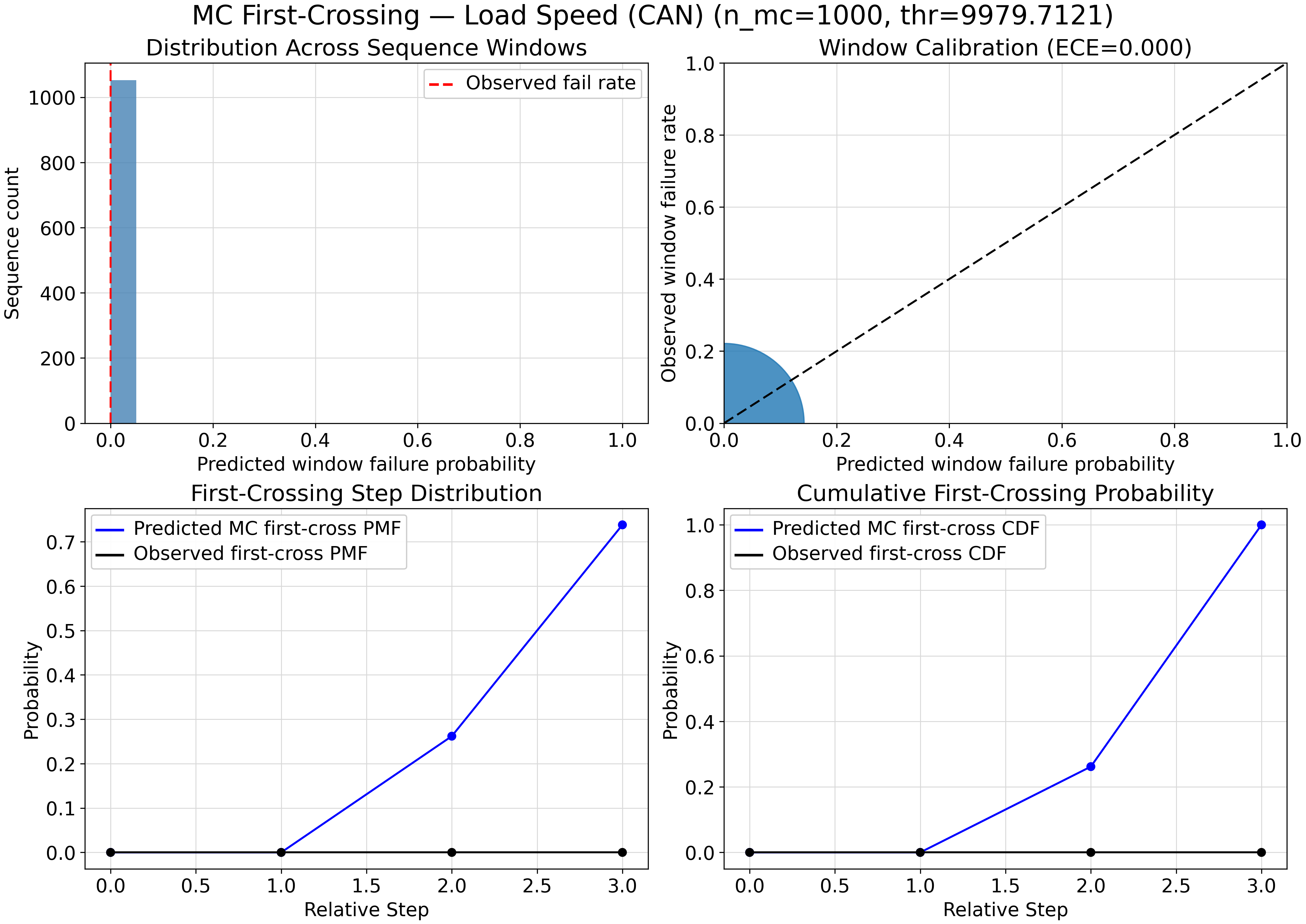}
\caption{Load Speed (CAN).}
\end{subfigure}
\caption{Monte Carlo first-crossing analysis for drive motor current and load speed (CAN).}
\label{fig:reliability_mc_representative}
\end{figure}

\begin{figure}
\centering
\begin{subfigure}[t]{0.49\textwidth}
\centering
\includegraphics[width=\linewidth]{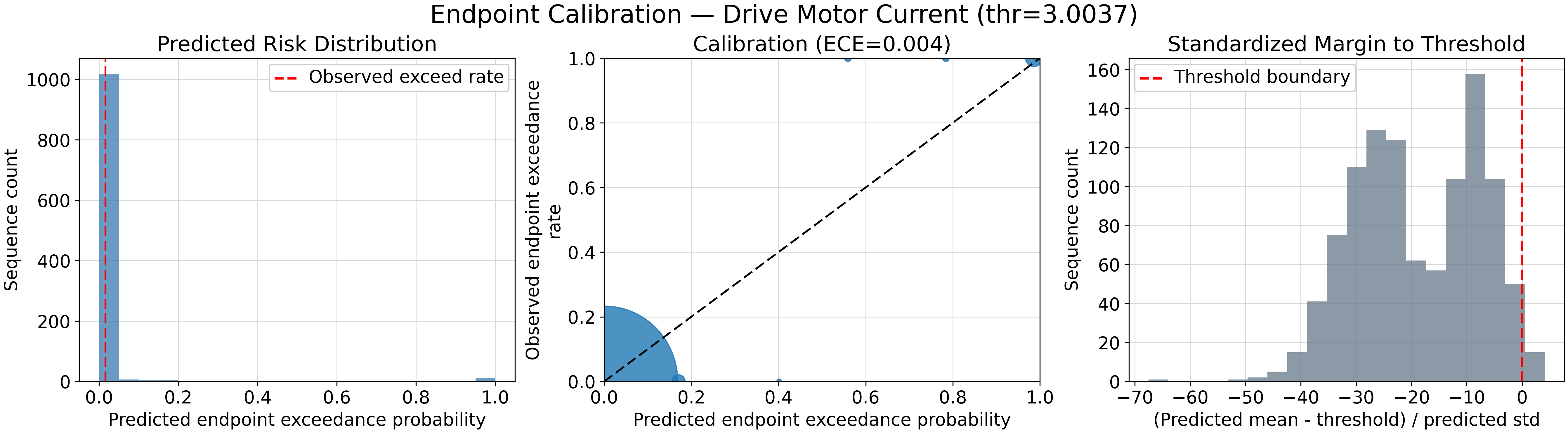}
\caption{Drive Motor Current.}
\end{subfigure}
\hfill
\begin{subfigure}[t]{0.49\textwidth}
\centering
\includegraphics[width=\linewidth]{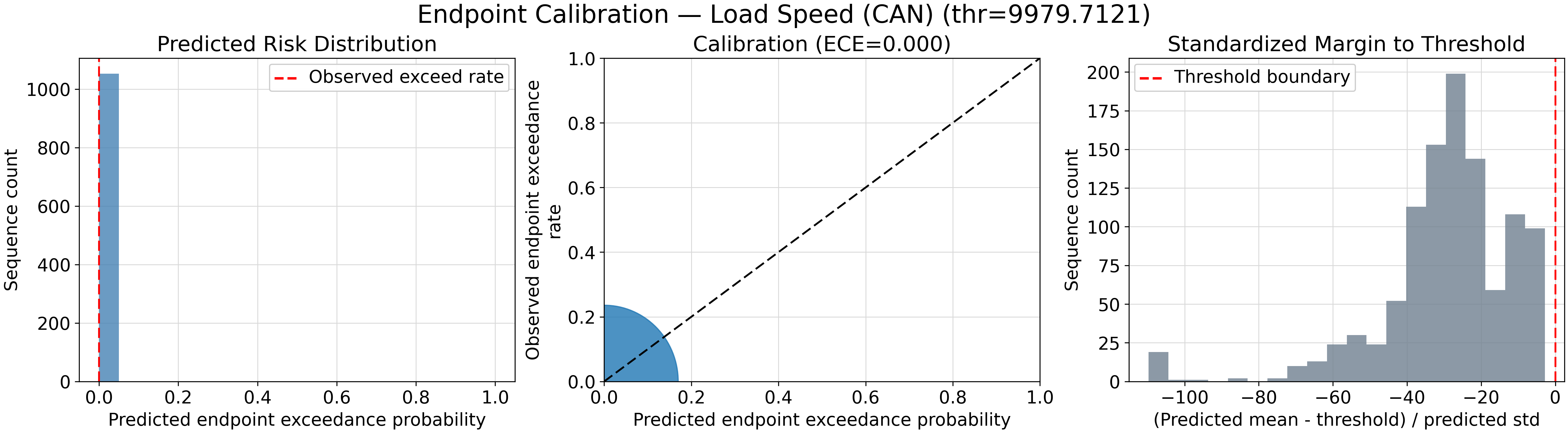}
\caption{Load Speed (CAN).}
\end{subfigure}
\caption{Endpoint calibration analysis for drive motor current and load speed (CAN).}
\label{fig:reliability_endpoint_representative}
\end{figure}

Figures~\ref{fig:reliability_sequence_representative}--\ref{fig:reliability_endpoint_representative} show the sequence-local exceedance, Monte Carlo first-crossing, and endpoint calibration results for drive motor current and, as a representative low-risk counterexample, load speed (CAN).

For drive motor current, the Weibull survival analysis yielded a fitted shape parameter \(\beta=1.7147\) and scale parameter \(\eta=19.3964\), corresponding to a mean life of 17.30 relative steps, a median life of 15.66 relative steps, and a B10 life of 5.22 relative steps. The observed fail fraction was 0.0646 and the predicted Monte Carlo fail fraction was 0.0600. The empirical median crossing was not reached within the held-out horizon. For all other outputs, including load speed (CAN), the Weibull fit was skipped because fewer than two observed failures were available.

\begin{figure}
\centering
\includegraphics[width=0.78\linewidth]{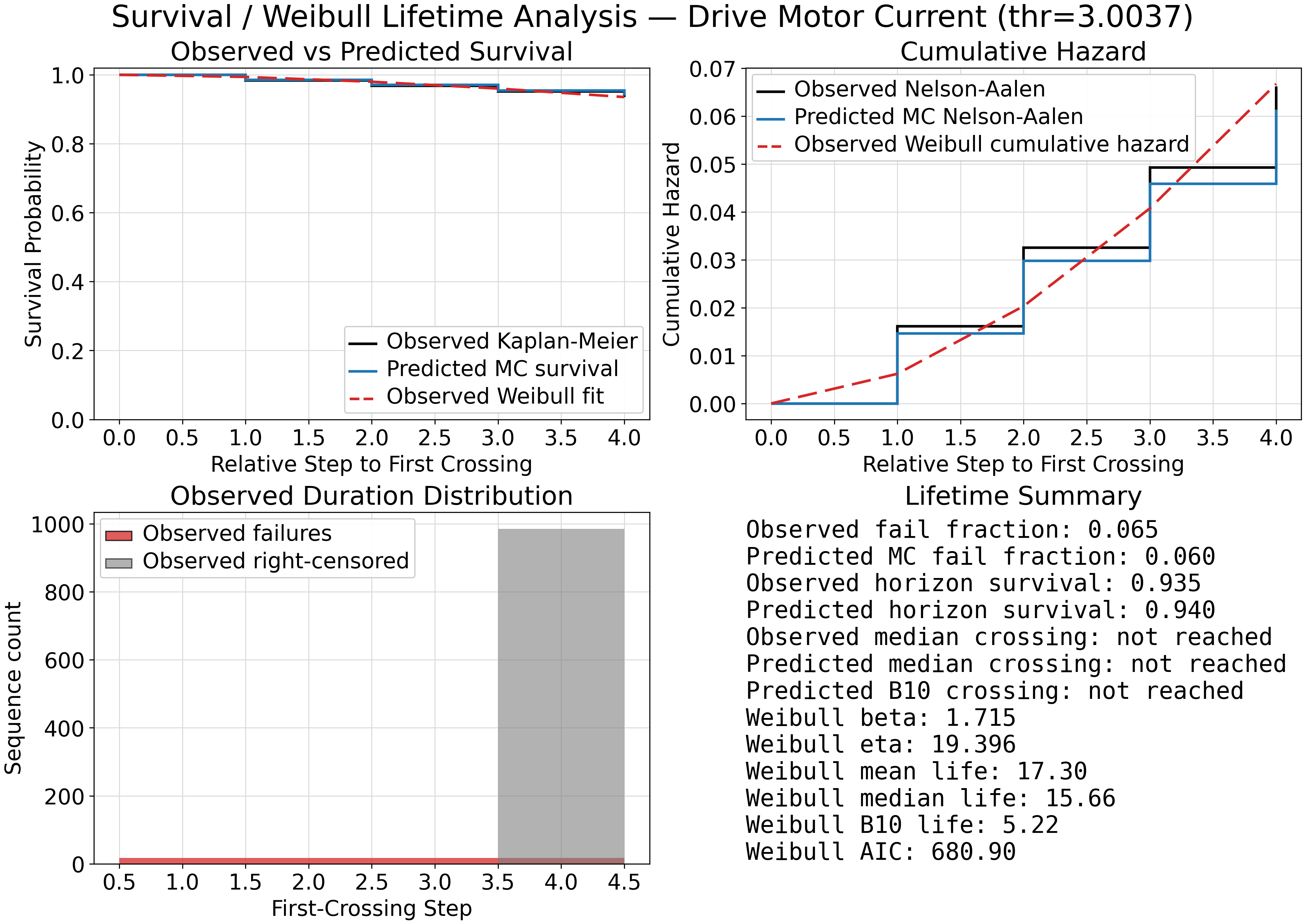}
\caption{Survival and Weibull lifetime analysis for drive motor current, the only output with sufficient observed failures for a stable fit on the held-out run.}
\label{fig:reliability_weibull_drive_motor_current}
\end{figure}

%%%%%%%%%%%%%%%%%%%%%%%%%%%%%%%%%%%%%%%%%%%%%%%%%%%%%%%%%%%%%%%%%%%%%%%%%%%%%%%%%%%%%%%%
\subsection{Component--Material Behavior Results}

\subsubsection{Fatigue Behavior of the Output Shaft}
\begin{figure}
    \centering    \includegraphics[width=0.55\textwidth]{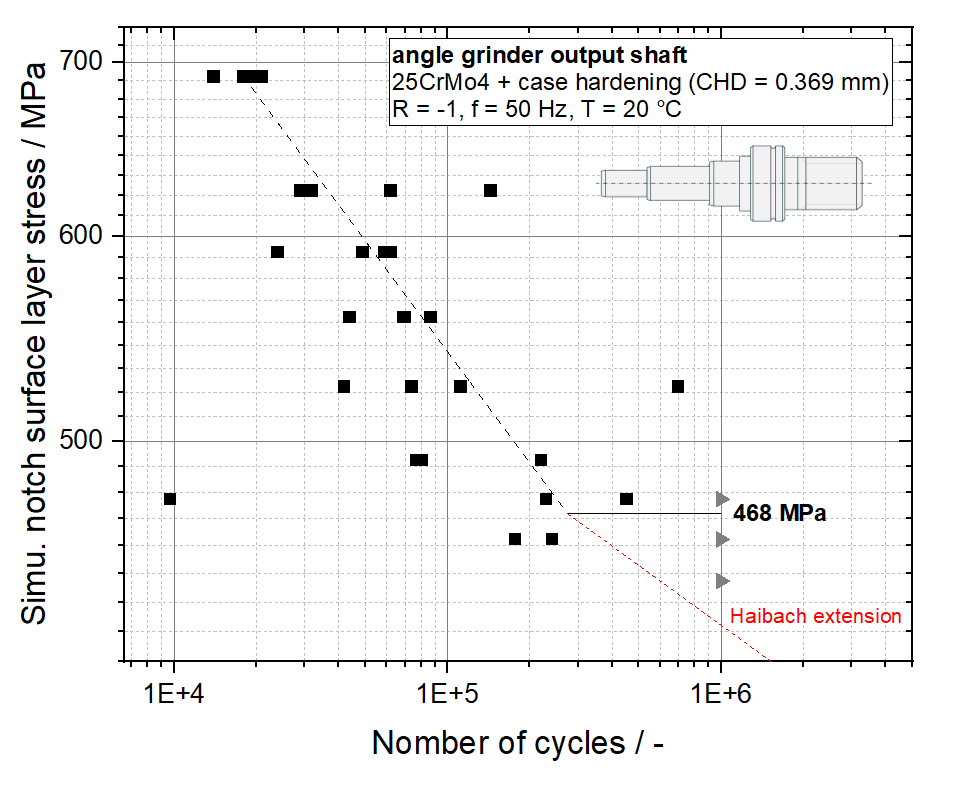}
    \caption{S--N curve of the output shaft under rotating bending, showing the Basquin fit to the finite-life data, the Haibach extension below the endurance limit, and the experimentally determined fatigue strength at 468~MPa.}
    \label{fig:sncurve}
\end{figure}

The S--N curve of the output shaft under rotating bending is shown in Fig.~\ref{fig:sncurve}. The experimental data exhibit a typical stress--life trend with noticeable scatter in the finite-life regime. A Basquin fit provides a consistent representation in the high-cycle range. The endurance limit was determined at 468~MPa by the staircase method; no failure was observed below this level within the investigated cycle range. For the subsequent variable-amplitude assessment, the S--N curve was extended using the Haibach modification to account for cycles below the endurance limit.

\subsection{Local stress state and material characterization}

Finite element analysis of the rotating bending configuration quantified the stress concentration at the notch root of the smallest shaft diameter, which also corresponds to the experimentally observed fracture location. The resulting local notch stress was therefore used as the fatigue-relevant stress quantity for the subsequent material assessment. For the service case, a separate simulation campaign was performed to establish an interpolation-based relation between external bending loads and local notch stress. This relation was used to reconstruct the time-resolved notch stress history from the measured force and bending-load records.

Hardness mapping revealed an effective case-hardened depth of approximately 0.369~mm according to ISO~2639. Fractographic observations showed surface-initiated fatigue cracking at the notch, followed by crack propagation through the hardened layer toward the lower-hardness core. This confirms that the near-surface region governs the fatigue response under rotating bending. Accordingly, the case-hardened depth was used as the upper crack-growth boundary in the Paris-law-based assessment.

\subsection{Crack propagation and reusability assessment}

To obtain a more damage-sensitive assessment, crack propagation was evaluated using a Paris-law-based model. The threshold-based initial crack length and the fracture-mechanics-based critical crack length were calculated first. However, the calculated initial crack size remained below the imposed microstructural lower bound, while the calculated critical crack length exceeded the effective case-hardened depth. Therefore, the crack-growth interval was bounded by an effective initial crack length of an initial crack length of \(a_0 = 4.4 \times 10^{-3}\,\mathrm{mm}\) and a critical crack length of \(a_{\mathrm{c}} = 0.369\,\mathrm{mm}\), corresponding to the case-hardened depth. The resulting crack-growth curves describe the number of cycles required for a crack to propagate from $a_0$ to $a_c$ under nominal and amplified service loading.

Under nominal service loading, the model predicts the highest reusability. Selective amplification of the upper 10\% of the stress-amplitude distribution strongly reduces the predicted crack propagation life. The estimated reusability decreases from 31 service phases for the original load spectrum to 20, 3, and 1 service phases for amplification factors of 1.2, 1.6, and 2, respectively. Each service phase corresponds to 400 operating hours at 9000~rpm. This demonstrates that the material-side RUL is governed primarily by rare high-load events rather than by the average load level.

The corresponding stress--time histories show that the amplification affects only the upper tail of the spectrum, while the overall temporal structure of the service load remains unchanged. This isolates the effect of high-amplitude events on crack propagation life.

\begin{figure}[]
    \centering
    \includegraphics[
        width=0.9\textwidth,
        trim=1.0cm 5cm 1.0cm 5cm,
        clip
    ]{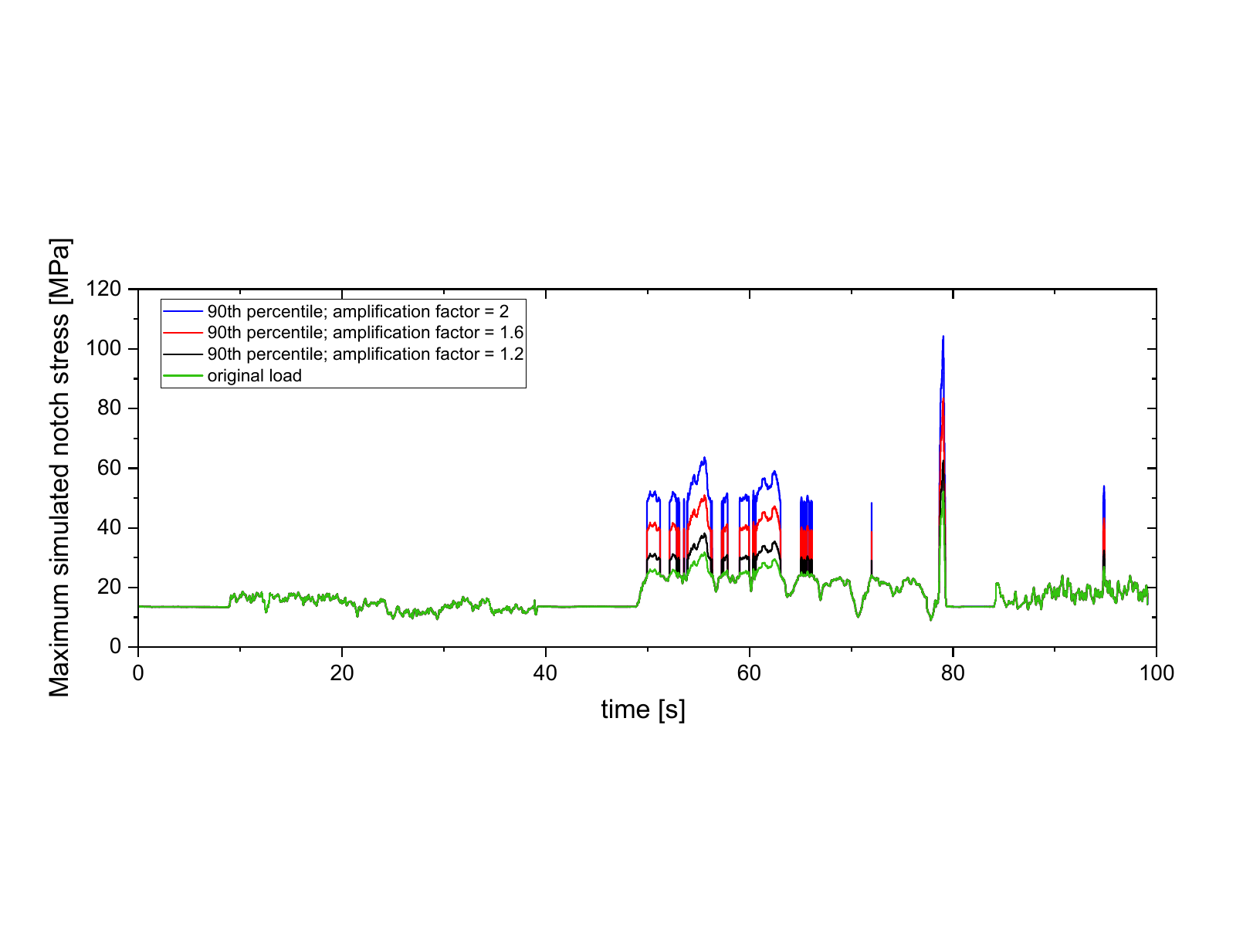}
    \caption{Predicted maximum notch stress--time history under nominal service loading and under selective amplification of stress values above the 90th percentile.}
    \label{fig:load-spectrum}
\end{figure}

\begin{figure}[]
    \centering
    \includegraphics[
        width=0.8\textwidth,
        trim=0.5cm 1cm 0.5cm 0.5cm,
        clip
    ]{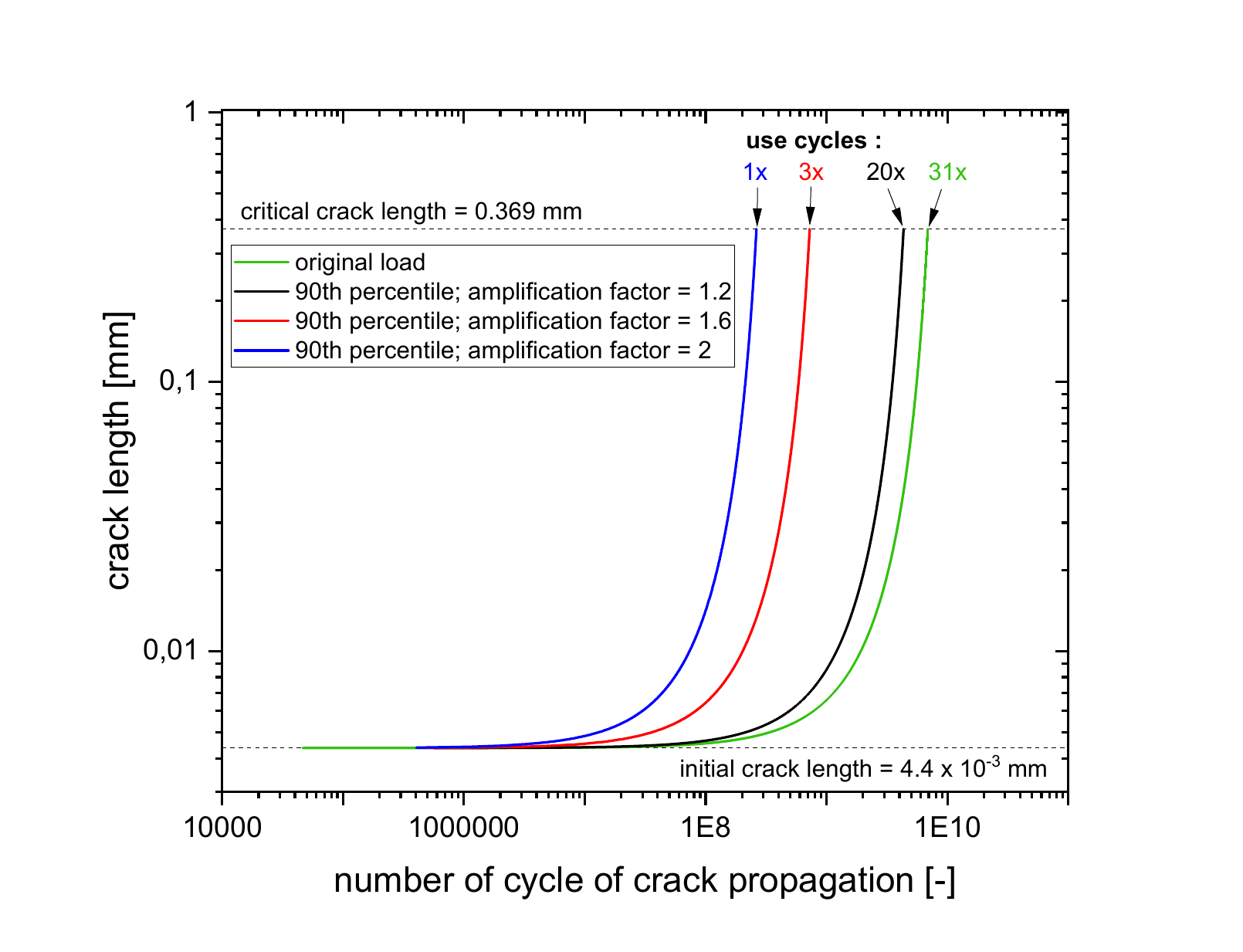}
    \caption{Predicted crack propagation life and derived reusability under nominal loading, selectively amplified loading with scaling factors of 1.2, 1.6
    , and 2.0 applied to the upper 10\% of the stress distribution ($\geq$90th percentile), and the grinding-only load profile. Reusability is expressed in 400~h service phases at 9000~rpm.}
    \label{fig:material-rul}
\end{figure}
%%%%%%%%%%%%%%%%%%%%%%%%%%%%%%%%%%%%%%%%%%%%%%%%%%%%%%%%%%%%%%%%%%%%%%%%%%%%%%%%%%%%%%%%
\subsection{Consolidation of Functional--Material Replay Results}
\begin{figure}
\centering
\includegraphics[width=0.65\linewidth]{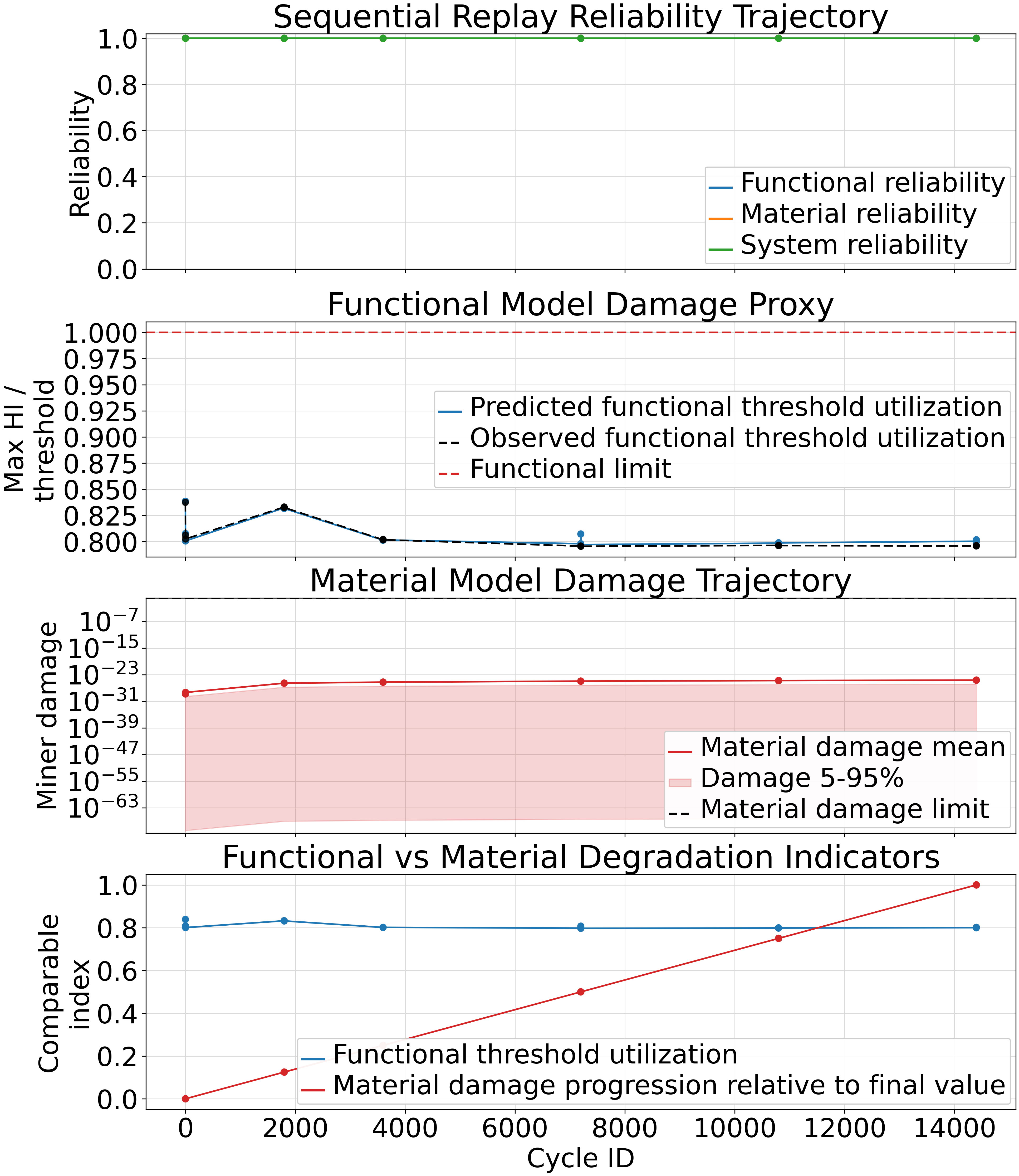}
\caption{Sequential replay results for Algorithm~\ref{alg:combined_reliability} over 18 ordered inspection files (cycle IDs 1--14400). Panel~1: functional reliability summary, material reliability, and system reliability indicator as defined in Section~4.3. Panel~2: predicted and observed maximum health-indicator threshold utilization across all outputs, with the functional failure limit at 1.0. Panel~3: cumulative Miner damage (log scale) with 5--95\% bootstrap uncertainty band and the material failure limit at \(D=1\). Panel~4: normalized functional threshold utilization and material damage progression for cross-domain comparison.}
\label{fig:sequential_replay_lstm}
\end{figure}

Algorithm~\ref{alg:combined_reliability} was evaluated by replaying the selected LSTM checkpoint over 18 ordered MAT files spanning cycle identifiers 1 to 14400. The same usage-history stream was processed simultaneously by the functional and material branches, and the resulting trajectories are shown in Fig.~\ref{fig:sequential_replay_lstm}.
At the inspection level, both branches remained within their respective assessed criteria. The final values at cycle 14400 were \(R_{\mathrm{func}}=1.0\), \(R_{\mathrm{mat}}=1.0\), and \(R_{\mathrm{sys}}=1.0\). Neither a predicted nor observed B10 or median failure cycle was reached across the 18 inspection points. The logged governing branch was material, reflecting a tie since both branches held at 1.0. This tie does not establish physical dominance of either degradation mechanism, and \(R_{\mathrm{sys}}=1.0\) concerns the modeled criteria rather than complete-machine survival probability.

Within the full window sequence, drive motor current was again the most critical output. Its mean sequence-window failure probability was 0.0932, compared with an observed window failure rate of 0.0921. The Monte Carlo first-crossing probability was 0.0931, the endpoint risk was 0.0253, and the predicted horizon survival probability was 0.9069. The fitted Weibull parameters were \(\beta=1.6948\) and \(\eta=15.8934\). Load speed (CAN) remained effectively risk-free, with a mean window failure probability of \(1.5 \times 10^{-5}\).

On the material side, each inspection file contributed 9 material stress levels from the load collective, all of which passed the validity threshold. The mean notch stress per file was approximately 2.88~MPa. The incremental Miner damage per file was approximately \(1.52 \times 10^{-29}\), accumulating to a cumulative value of approximately \(2.19 \times 10^{-25}\) across all 18 files — negligible relative to the failure threshold of 1.0. Consequently, \(R_{\mathrm{mat}}\) remained at 1.0 throughout the full replay. The extremely small damage values indicate that the service stress amplitudes of approximately 2.88~MPa are well below the shaft's 468~MPa endurance limit, placing the component well within the safe fatigue regime under the measured load conditions.
%%%%%%%%%%%%%%%%%%%%%%%%%%%%%%%%%%%%%%%%%%%%%%%%%%%%%%%%%%%%%%%%%%%%%%%%%%%%%%%%%%%%%
%%%%%%%%%%%%%%%%%%%%%%%%%%%%%%%%%%%%%%%%%%%%%%%%%%%%%%%%%%%%%%%%%%%%%%%%%%%%%%%%%%%%%
\section{Discussion}\label{sec:discussion}
%\subsection{Answering the Research Question}
This study investigated how the future functional behavior and material-fatigue states of a returned angle grinder can be jointly assessed from current condition and usage history to support Circular Factory redeployment decisions. The results show that this is feasible through two linked reliability branches: a system-level functional behavior model that predicts future function fulfillment from state and force--torque history, and a component-level material model that translates load history into shaft stress, fatigue damage, and crack-growth-based reusability. The sequential replay then consolidates both branches into functional and material reliability estimates and a system reliability indicator identifying the limiting assessed requirement.

The research question \textbf{\emph{
How can the current state, usage history, and future usage scenario of a returned product be represented in a unified functional–material reliability
space to support instance-specific redeployment decisions in the CF under uncertainty?}} is therefore answered as follows: the framework demonstrates a consistent method for joint functional--material assessment, but the present case mainly proves functional-risk estimation and workflow integration. It does not yet provide strong evidence of a physical coupling between functional degradation and material fatigue, because the investigated shaft loads remained far below the fatigue-critical regime.

\subsection{Functional Evidence for Redeployment Risk Assessment}
The functional results show that future system behavior can be predicted with high accuracy from the selected state and usage-history variables. The LSTM achieved a mean 2\%-tolerance accuracy of 0.9652, mean NRMSE of 0.0297, and mean \(R^2\) of 0.8365 across nine outputs. The thermal variables were predicted almost perfectly, with \(R^2 = 0.9999\), indicating that the current state and short load history capture the dominant thermal behavior.

The more relevant redeployment insight comes from the dynamic outputs. Drive motor current and load speed were the most challenging variables, with 2\%-tolerance accuracies of 0.9174 and 0.8675, but high \(R^2\) values of 0.9750 and 0.9924. This means the model captures their global trends well, while local deviations remain important for threshold-based functional decisions.

The input ablation confirms that usage-history design is critical. Adding measuring-shaft torque substantially improved the prediction of drive motor current and load speed, showing that Circular Factory assessment should not rely on static inspection data alone. Physically relevant load-history variables are required to distinguish products with similar current states but different future risks.

The LSTM outperformed GRU and xLSTM in this setting. This should not be overstated as general LSTM superiority. Rather, for short overlapping usage windows with local patterns already extracted by the convolutional encoder, a conventional LSTM provides sufficient memory capacity, stable training, and lower architectural complexity.

From a reliability perspective, drive motor current was the only output with enough exceedance events for meaningful calibration. Its predicted window failure probability was 0.0600 compared with an observed rate of 0.0646, and its endpoint risk was 0.0154 compared with an observed rate of 0.0161. The low ECE values, 0.0085--0.0086 for window-level risk and 0.0033 for endpoint risk, indicate that the uncertainty estimates provide useful functional risk information. For the other outputs, the results show stable prediction in safe regimes, but not yet robust lifetime calibration.

\subsection{Material Reusability and Load-History Sensitivity}

The component--material assessment indicates a high reuse potential of the output shaft under the investigated nominal service loading. The experimentally determined fatigue strength is 468~MPa, whereas the maximum reconstructed local notch stress under nominal service conditions is approximately 50~MPa. Accordingly, the S--N/Miner evaluation predicts negligible cumulative damage for an initially intact shaft, indicating that the component is not fatigue-limited under the measured load spectrum.

The Paris-law-based analysis addresses a more conservative damage scenario by assuming the presence of an initial crack-like defect. Under nominal service loading, the calculated crack propagation life is approximately $6.86 \times 10^{9}$ fatigue cycles, corresponding to about 31 potential service phases of 400 operating hours each at 9000~rpm, based on the manufacturer-specified service life. However, this reuse potential is highly sensitive to the upper tail of the load spectrum. Increasing the upper 10\% of stress amplitudes by factors of 1.2 and 1.6 reduces the predicted reuse potential to approximately 20 and 3 service phases, respectively. For a factor of 2, only one service phase remains.

This is a central implication for the Circular Factory: component reusability cannot be inferred from operating time alone. Rare high-load events and overload history can dominate crack propagation life and therefore determine whether direct reuse is still justified. Reliable component-level reuse planning therefore requires recorded or reconstructed load histories, not only inspection data at the time of return. 
%\hl{Related inspections of used angle-grinder spindle shafts identified localized surface defects, including scratches and occasional pits} \citep{HemmerichKochMas2026_1000197183}. \hl{Such defects may act as local stress raisers and promote fatigue-crack initiation under cyclic loading, however, this effect was not quantified in the present study.} 
Comparing the predicted remaining fatigue capability with the requirements of the intended next use provides an additional criterion for deciding whether direct reuse remains appropriate.

\subsection{Integrated Functional--Material Replay}
The sequential replay across 18 ordered inspection files demonstrates that the functional and material branches can be integrated coherently within a single streaming reliability workflow. The functional reliability summary, material reliability, and system reliability indicator remained at \(R_{\mathrm{func}} = 1.0\), \(R_{\mathrm{mat}} = 1.0\), and \(R_{\mathrm{sys}} = 1.0\). Drive motor current again appeared as the dominant functional risk variable, with a predicted window failure probability of 0.0932 compared with an observed rate of 0.0921.

The material branch produced valid damage updates, but the accumulated Miner damage remained only approximately \(2.19 \times 10^{-25}\), far below the failure threshold of 1.0. Thus, the replay validates algorithmic integration and consistency, but not a strong material degradation trajectory. Under the tested conditions, redeployment risk is more likely governed by functional behavior than by output-shaft fatigue.

\subsection{Implications for Circular Factory Decision-Making}
The framework supports a shift from inspection-based classification to prediction-based redeployment. A returned product should not only be judged by whether it currently operates, but by whether it is expected to fulfill its function in the next use phase and whether its critical components retain sufficient structural capability.

Following product return, the framework operates at inspection, using the measured current state and available usage history to inform functional and material assessment and the system reliability indicator. The crack-growth assessment additionally provides scenario-specific component reuse potential. These outputs support routing decisions alongside product requirements and available processing resources. Where intervention is required, recovered components can be reconditioned or replaced and matched with suitable recovered or newly manufactured parts before assembly and system verification. Products suitable for direct or restricted reuse proceed to verification, while rejected items follow material-recovery or recycling routes. Verified products are redeployed and may return after a subsequent service phase. Detailed production scheduling and factory IT implementation remain outside the present implementation.

\begin{figure}[]
    \centering
    \includegraphics[
        width=0.95\textwidth,
        trim=0.0cm 0.0cm 0.0cm 0.0cm,
        clip
    ]{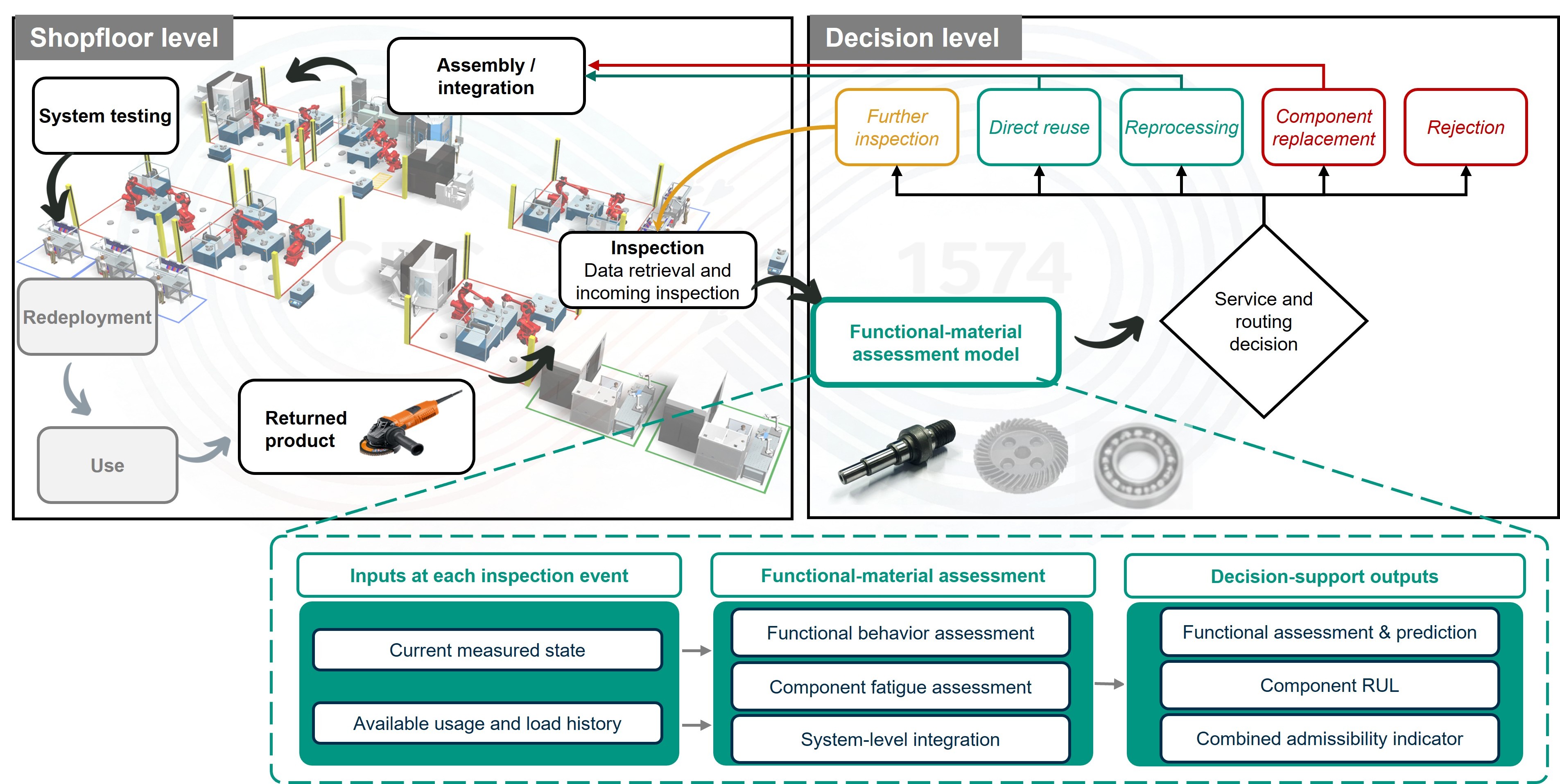}
        \caption{Functional-material assessment in the Circular Factory return-to-redeployment process. At each inspection event, the current state and available usage and load history are used to update the functional assessment, component RUL, and combined indicator \(R_{\mathrm{Sys}}\). These outputs support the subsequent routing decision.}
    \label{fig:IntegrationInCF}
\end{figure}

The results also show that product-level and component-level decisions may differ. An angle grinder may become functionally unsuitable while the output shaft remains reusable. Conversely, a functional tool may still require component inspection if material risk is high. Figure~\ref{fig:IntegrationInCF} positions the proposed framework at the inspection and service-decision level and shows how current-state and usage-history information support routing between direct reuse, restricted reuse, further inspection, reprocessing, component replacement, and rejection or recycling. Because the output shaft transmits torque, its structural failure provides a sufficient path to loss of drivetrain function. Nevertheless, fatigue is one decision criterion rather than the sole determinant of Circular Factory operations. This supports more differentiated Circular Factory decisions such as direct reuse, restricted reuse, reconditioning, component replacement, further inspection, or rejection.

Fatigue damage can accumulate over successive service phases while current functional behavior remains acceptable. Its assessment therefore contributes information beyond present operability. The nominal results indicate substantial output-shaft reuse potential, supporting consideration of recovered shafts for further service alongside functional performance and other acceptance criteria. For Circular Factory operations, this information helps determine which recovered components remain candidates for reuse, which require further inspection or intervention, and which require replacement before assembly and redeployment. These decisions inform product routing, component matching, and replacement-part requirements. Quantitative effects on factory throughput, cost, and resource utilization remain to be evaluated.
For research, the contribution is the integration of data-driven functional prognosis and material-informed fatigue assessment in a redeployment-oriented reliability workflow. This connects PHM, reliability engineering, and circular manufacturing in a way that supports instance-specific decisions rather than population-level lifetime assumptions. The framework provides decision support only; detailed route selection, production scheduling, and factory IT implementation remain outside the present scope.

\subsection{Limitations and Research Outlook}
The functional model was evaluated on controlled test-bench data with a fixed load cycle. Generalization to real Circular Factory conditions still requires validation on multiple returned products, heterogeneous usage histories, variable operating profiles, and reprocessed product states.

Future work will investigate concurrent degradation and diverse failure root causes through dedicated experiments. The resulting data will support evaluation of the end-to-end functional predictor beyond the present operating conditions and provide a basis for enriching the observed state representation, incorporating additional failure models, and investigating physical interactions and statistical dependencies. Predictive performance under concurrent component-specific degradation remains to be established through these investigations.

The reliability analysis is also limited by the availability of events. Only the drive motor current had sufficient exceedance events for meaningful calibration. More event-rich degradation data are needed to validate functional reliability estimates across multiple indicators.

The material model is load-case-specific. The S--N/Miner assessment assumes linear damage accumulation and neglects load sequence effects, while the Paris-law model assumes long-crack propagation and uses literature-based crack-growth parameters. Stress reconstruction also relies on simplified finite-element boundary conditions. Therefore, the predicted reuse potential should be interpreted as a scenario-specific estimate rather than a universal lifetime value for all shafts.

Future work should focus on validating the functional model under heterogeneous operating conditions, calibrating shaft-specific crack-growth parameters, improving load-to-stress reconstruction, collecting fatigue-relevant service histories, and testing scenario-conditioned predictions against prescribed future load profiles. These steps are necessary to move from an integrated prototype toward a robust Circular Factory decision-support method.
%%%%%%%%%%%%%%%%%%%%%%%%%%%%%%%%%%%%%%%%%%%%%%%%%%%%%%%%%%%%%%%%%%%%%%%%%%%%%%%%%%%%%%%%%
%%%%%%%%%%%%%%%%%%%%%%%%%%%%%%%%%%%%%%%%%%%%%%%%%%%%%%%%%%%%%%%%%%%%%%%%%%%%%%%%%%%%%%%%%
\section{Conclusion}\label{}
This paper presented an uncertainty-aware framework for instance-specific assessment of returned angle grinders in a circular-factory context. The framework combines two complementary branches: a conditional sequence model that predicts future functional behavior from the current tool state and recent force--torque history, and a material-fatigue model that translates the same operational loading into cumulative shaft damage and a corresponding reliability estimate. Together, they are integrated into a streaming algorithm that updates functional and material reliability and a system reliability indicator at each inspection point while accounting for gaps between non-consecutive observations.

On the functional side, the conventional LSTM achieved the best trade-off among the tested recurrent backbones, with a mean held-out \(2\%\)-tolerance accuracy of 0.9652 and mean \(R^2\) of 0.8365 across nine outputs, outperforming both GRU and xLSTM. Thermal variables were predicted near-perfectly; drive motor current and load speed were the most challenging, yet their global trajectories were captured well (\(R^2 = 0.9750\) and \(0.9924\)). The uncertainty estimates proved actionable: for drive motor current, the only output with observable exceedance events, predicted and observed failure probabilities agreed closely across all four reliability views, with window ECE below 0.01 and Weibull parameters \(\beta = 1.7147\), \(\eta = 19.3964\), yielding a B10 life of 5.22 relative steps. On the material side, the Paris-law crack propagation analysis yielded a propagation life of approximately \(6.86 \times 10^{9}\) fatigue cycles, corresponding to approximately 31 potential service phases of the output shaft at nominal service loading. This lifetime was found to be sensitive to high-load events: amplifying the upper 10th percentile of stress amplitudes by a factor of 1.6 reduced the predicted number of service phases to approximately three.

The sequential replay over 18 ordered inspection files demonstrated that the functional and material assessment streams operate coherently from the same usage-history data, with $R_{\mathrm{func}}$, $R_{\mathrm{mat}}$, and $R_{\mathrm{sys}}$ remaining stable at 1.0 across the full horizon. Functional criticality rankings were preserved, with drive motor current again yielding the highest exceedance probability and Weibull statistics consistent with the held-out analysis. On the material side, valid damage windows were generated, but the cumulative Miner damage remained negligible relative to the failure threshold because the reconstructed service stresses were far below the experimentally determined endurance limit of 468~MPa. The result therefore verifies algorithmic integration and internal consistency, but does not yet demonstrate progressive material degradation under fatigue-critical loading. The system reliability indicator identifies the limiting modeled requirement, while additional failure mechanisms and their interactions remain an extension of the framework.

For the crack-propagation-based assessment, the current results indicate a nominal reusability of approximately 31 service phases. Selective amplification of the upper 10\% of the stress-amplitude distribution reduces the predicted reuse potential to approximately 20, 3, and 1 service phases for amplification factors of 1.2, 1.6, and 2, respectively. This confirms that material-side RUL is highly sensitive to rare high-load events. Future work should therefore extend the evaluation to more severe and heterogeneous usage histories, reprocessed product instances, experimentally crack-growth parameters, and scenario-conditioned future load profiles.
%%%%%%%%%%%%%%%%%%%%%%%%%%%%%%%%%%%%%%%%%%%%%%%%%%%%%%%%%%%%%%%%%%%%%%%%%%%%%%%%%%%%%
%%%%%%%%%%%%%%%%%%%%%%%%%%%%%%%%%%%%%%%%%%%%%%%%%%%%%%%%%%%%%%%%%%%%%%%%%%%%%%%%%%%%%

\printcredits
\par\medskip
\textbf{Declaration of generative AI and AI-assisted technologies in the manuscript preparation process}\\
During the preparation of this work, the authors used DeepL and ChatGPT (OpenAI) in order to improve the structure, readability, and phrasing of the manuscript. 
After using these tools, the authors reviewed and edited the content as needed and take full responsibility for the publication's content.\\
\\
\textbf{Declaration of competing interest} \\
The authors declare that they have no known competing financial interests or personal relationships that could have appeared to influence the work reported in this paper.\\
\\
\textbf{Acknowledgments} \\
Funded by the Deutsche Forschungsgemeinschaft (DFG, German Research Foundation) - SFB 1574 - 471687386.

%\section{}\label{}
\bibliographystyle{cas-model2-names}

\bibliography{cas-refs}
% \bibliography{Main/RUL-Crack}



\end{document}